\documentclass[10pt]{article}

\usepackage[preprint]{tmlr}

\usepackage[utf8]{inputenc}
\usepackage[table]{xcolor}
\usepackage{graphicx}
\usepackage[percent]{overpic}
\definecolor{trajectorybg}{RGB}{235,240,246}
\usepackage{booktabs}
\usepackage{amsmath}
\usepackage{amssymb}
\usepackage{amsfonts}
\usepackage{mathtools}
\usepackage{microtype}
\usepackage{wrapfig}
\usepackage{etoolbox}
\BeforeBeginEnvironment{wrapfigure}{\setlength{\intextsep}{0pt}}
\usepackage{multirow}
\usepackage{pifont}
\usepackage{array}
\usepackage{url}
\usepackage{hyperref}
\hypersetup{hidelinks}

\newcommand{\eg}{\textit{e.g.}}

\newcommand{\upx}[1]{\textcolor{violet}{\footnotesize \ $\uparrow${#1}}}
\newcommand{\downx}[1]{\textcolor{red!20!green!20!blue!50}{\footnotesize \ $\downarrow${#1}}}
\DeclareRobustCommand{\rev}[1]{{\color{black}#1}}
\newcommand{\TGnewrefs}{}
\forcsvlist{\listadd{\TGnewrefs}}{wu2024sld,yang2024idea2img,wang2024genartist,li2025reflectdit,zhuo2025reflectionflow}
\newcommand{\TGrevisionbibitem}[2][]{\color{black}\TGoriginalbibitem[#1]{#2}\ifinlist{#2}{\TGnewrefs}{\color{black}}{}}
\AtBeginEnvironment{thebibliography}{\let\TGoriginalbibitem\bibitem\let\bibitem\TGrevisionbibitem}
\newcommand{\cmark}{\ding{51}}
\newcommand{\xmark}{\ding{55}}

\title{T-GUI: Self-Correction in Unified Multimodal Model via Test-time Generation--Understanding Interaction}

\author{\name Yuanhuiyi Lyu \email ylyu650@connect.hkust-gz.edu.cn\\
  \addr The Hong Kong University of Science and Technology (Guangzhou)
\AND
\name Chi Kit Wong \email ckwong627@connect.hkust-gz.edu.cn\\
  \addr The Hong Kong University of Science and Technology (Guangzhou)
\AND
\name Chenfei Liao \email cliao127@connect.hkust-gz.edu.cn\\
  \addr The Hong Kong University of Science and Technology (Guangzhou)
\AND
\name Lutao Jiang \email ljiang553@connect.hkust-gz.edu.cn\\
  \addr The Hong Kong University of Science and Technology (Guangzhou)
\AND
\name Xu Zheng \email zhengx@gbu.edu.cn\\
  \addr Great Bay University
\AND
\name Zexin Lu \email zexin.lu@connect.polyu.hk\\
  \addr The Hong Kong Polytechnic University
\AND
\name Linfeng Zhang \email zhanglinfeng@sjtu.edu.cn\\
  \addr Shanghai Jiao Tong University
\AND
\name Xuming Hu\thanks{Corresponding author.} \email xuminghu@hkust-gz.edu.cn\\
  \addr The Hong Kong University of Science and Technology (Guangzhou)
}

\def\month{12}
\def\year{2026}
\def\openreview{\url{https://openreview.net/forum?id=CMHtHIjKSV}}

\begin{document}

\maketitle

\begin{abstract}
Recent works have made notable advancements in enhancing the image generation of the Unified multimodal model (UMM) at test time.
\rev{However, prompt-based methods typically factorize the pipeline into ``\textit{understanding-then-generation}", with understanding limited to prompt refinement before image generation.}
In this work, we propose the Test-time Generation-Understanding Interaction (T-GUI), a self-correction framework that harnesses the robust understanding capabilities of the UMM to reinforce its performance in image generation.
The core idea is to iteratively incorporate understanding-driven guidance and self-verification into the generation process.
To achieve this, we introduce ``\textit{Image Editing}" as a bridge to infuse understanding into the generation process. Initially, we verify the generated image and incorporate the understanding of the UMM into the editing instructions. We then use these instructions to edit the generated image.
\rev{Our T-GUI shows an overall score improvement of \textbf{1.45 points} with short prompts and \textbf{4.57 points} with long prompts on TIIF-Bench relative to the same BAGEL backbone.}
\end{abstract}

\begin{figure}[t!]
    \centering
    \includegraphics[width=\linewidth]{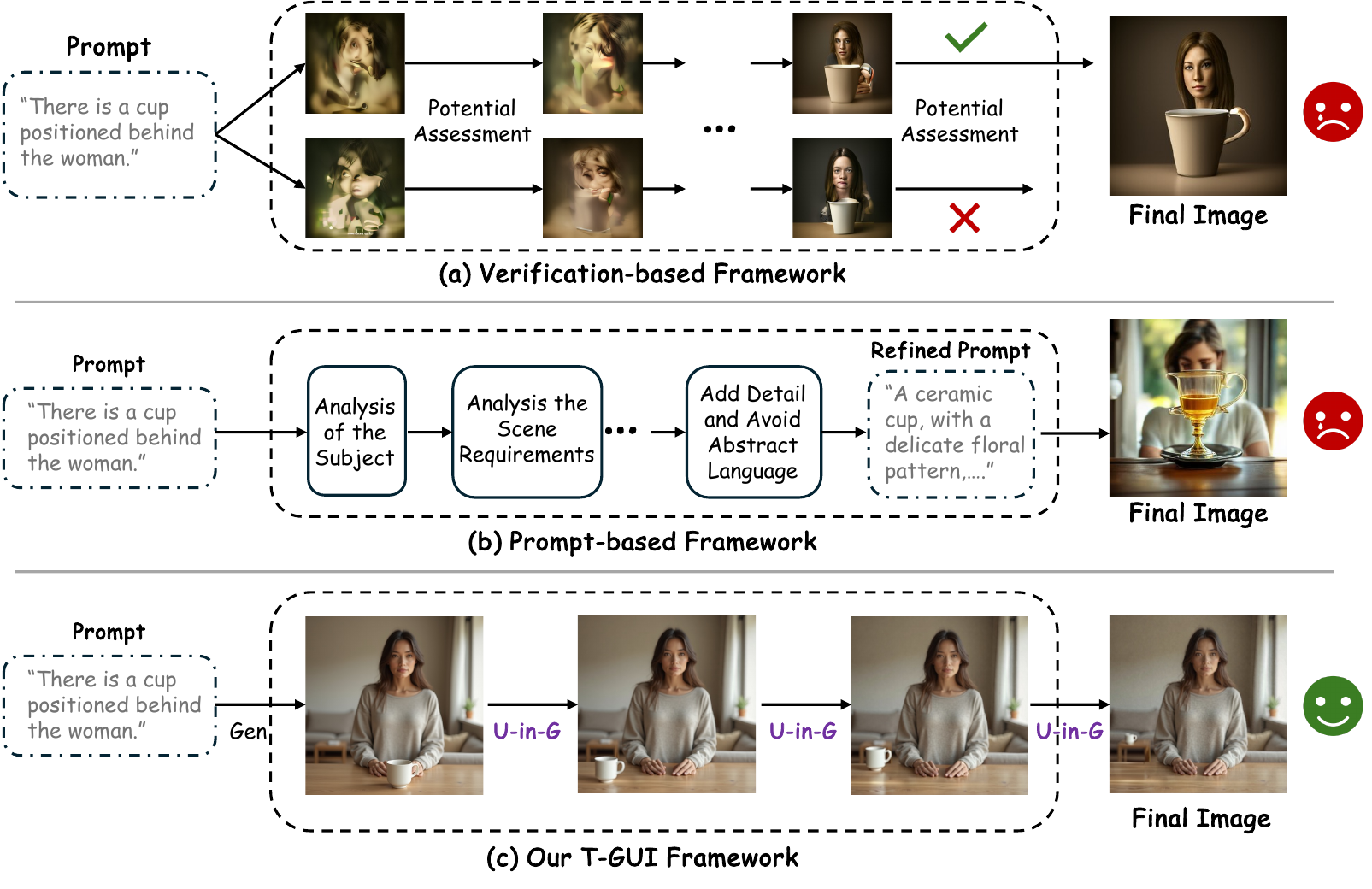}
    \vspace{-12pt}
    \caption{The overall of \textbf{(a)} Verification-based framework, \textbf{(b)} Prompt-based framework, and \textbf{(c)} Our T-GUI framework.}
    \vspace{-18pt}
    \label{fig: teaser}
\end{figure}

\section{Introduction}
Recent test-time augmented methods for the UMM have achieved notable progress in image generation. There are two primary categories of these methods: \textbf{(1)} \textbf{Verification-based framework}, which verifies and selects the generated images, and \textbf{(2)} \textbf{Prompt-based framework}, which enhances the input prompts through CoT.
However, both separate the understanding from the actual generation process, making it difficult to provide direct, fine-grained guidance during image synthesis.

First, for the verification-based framework (\eg, ImageCoT~\cite{zhang2025let}), as illustrated in Figure~\ref{fig: teaser} \textbf{(a)}, this method constructs multiple generative branches through repeated sampling and assesses the generative potential throughout the generation process. Specifically, the state of each branch is evaluated to determine whether it remains capable of generating. If the assessment gives a negative result, the branch is terminated.
Here, the understanding of the UMM is applied exclusively to the verification of intermediate images, selecting the best output from numerous samples. \rev{However, in this framework, the understanding capability is used for validation and filtering, with its guidance limited to selecting among sampled candidates.} As a result, the generative ability remains confined to scopes that can be reached through repeated sampling. As shown in Figure~\ref{fig: teaser} \textbf{(a)}, given the input prompt, {\fontfamily{qcr}\selectfont["There is a cup positioned behind the woman"]}, all outputs consistently place the cup in front of the woman. Consequently, even the best result obtained through the verification-based framework fails to satisfy the spatial relationship.

For the prompt-based framework, as shown in Figure~\ref{fig: teaser} \textbf{(b)}, CoT is employed to analyze the original prompt from multiple perspectives, \eg, the subject, scene requirements, and other relevant aspects, to derive a more refined prompt. This framework fully leverages the understanding capabilities of the UMM to enhance the initial prompt, and the resulting refined prompt is then used as the final input for image generation. This approach emphasizes the ``\textit{understanding-then-generation}" pipeline, in which the understanding phase precedes the generation process. \rev{However, this pipeline is limited to reasoning about the prompt before generation, while T-GUI additionally understands the generated image.} As illustrated in Figure~\ref{fig: teaser} \textbf{(b)}, although a refined prompt is obtained, the process does not engage with generation, thereby preventing the recognition of generative weaknesses, \eg, the spatial relationship between the cup and the woman. As a result, the refined prompt fails to guide the generation to the correct direction.

\rev{Beyond these two categories, iterative feedback is also explored in SLD~\citep{wu2024sld}, Idea2Img~\citep{yang2024idea2img}, GenArtist~\citep{wang2024genartist}, Reflect-DiT~\citep{li2025reflectdit}, and ReflectionFlow~\citep{zhuo2025reflectionflow}. 
T-GUI follows this iterative refinement principle, with a focus on using a single existing UMM for generation, verification, and editing, without relying on an external critic, detector, or tool library and without additional training.
We discuss these distinctions in Section~\ref{diff_with_existing_loop}.}

In this paper, we introduce Test-time Generation-Understanding Interaction (T-GUI), a test-time self-correction framework that harnesses the robust understanding capabilities of the UMM to reinforce their image generation performance at test time.
The \textit{\textbf{key insight}} of our T-GUI is to \textit{iteratively incorporate understanding-driven guidance and self-verification into the generation process.}

Specifically, as illustrated in Figure~\ref{fig: teaser} \textbf{(c)}, we first generate an initial image from the original prompt, which shows the generative ability of the UMM. We then evaluate the generated image and recognize its weaknesses, utilizing the understanding capabilities of the UMM. This result is subsequently integrated into the editing prompt, which carries the robust understanding capabilities of the UMM. Finally, we utilize the edited prompt to infuse this understanding into the generation process, guiding the generative direction and overcoming the limitations of the initial image.
Through this interleaved correction process, T-GUI mitigates generative limitations that are difficult to resolve by the verification-based or prompt-based framework.

\rev{Our T-GUI demonstrates overall gains of 1.45 and 4.57 points over the same BAGEL backbone under the short and long prompt settings of TIIF-Bench, respectively. On WISE, it improves the score from 0.52 to 0.72 without native thinking and from 0.70 to 0.74 when combined with native thinking.}
Our key contributions are summarized as follows:
\textbf{1)} We propose Test-time Generation-Understanding Interaction (T-GUI), a self-correction framework that reinforces the generation via the Generation-Understanding interaction at inference time;
\textbf{2)} We demonstrate the promising potential of the UMM to enhance the generative capabilities with the guidance from the strong understanding capabilities during inference;
\rev{\textbf{3)} Our T-GUI demonstrates overall gains of 1.45 and 4.57 points over the same BAGEL backbone in the short and long prompt settings of TIIF-Bench, respectively.}

\section{Related Work}


\subsection{Unified Multimodal Model}
Recent multimodal large language models~\citep{driess2023palm,peng2023kosmos,alayrac2022flamingo,zhu2023minigpt,wang2024qwen2,bai2025qwen2,chen2024internvl,li2023blip,lin2023video,gao2023llama,zhang2023llama} have demonstrated remarkable progress in visual understanding powered by the strong vision encoders~\citep{dosovitskiy2020image,radford2021learning,girdhar2023imagebind,zhu2023languagebind,guo2023point,lyu2024unibind}. Building on this success, unified visual understanding and generation within a single framework has emerged as a central research frontier. Early unified models, such as Transfusion~\citep{zhou2024transfusion}, Chameleon~\citep{team2024chameleon}, Emu3~\citep{wang2024emu3}, and Show-o~\citep{xie2024show}, employ visual tokenizers, \eg, VQ-VAE~\citep{van2017neural}, to encode images into sequences of tokens, to enable seamless multimodal understanding and generation further. However, these discrete visual tokens introduce visual information loss, limiting the extraction of fine-grained semantic content. To address this, the Janus~\citep{chen2025janus} series decouples visual encoding for understanding and generation by adopting separate encoders, though task conflicts within the shared parameter space can hinder its performance. Meanwhile, BAGEL~\citep{deng2025emerging} adopts a MoT architecture, assigning autoregressive text generation and diffusion-based image synthesis to distinct components. 
Despite their effectiveness, these models still struggle to fully exploit their strong understanding capabilities during the generative process, which restricts their ability to produce images with complex logical content.

\subsection{Test-time Augmentation for Image Generation}
\label{diff_with_existing_loop}
Test-time CoT~\citep{wei2022chain} has played an effective role in LLM~\citep{xia2024beyond,zhang2024chain,deng2024explicit,chen2025towards,kang2025c3ot,wu2025more,wang2024chain} and MLLM~\citep{wang2025multimodal,xu2024llava,shao2024visual,zhao2025cot,ma2025audio,jiang2025mme}, enabling them to decompose complex tasks into structured intermediate steps. Recent research has extended the test-time augmented framework into text-to-image generation, primarily through verification-based and prompt-based approaches. Verification-based methods, \eg, ImageCoT~\citep{zhang2025let}, generate multiple candidate images via repeated sampling and then apply CoT to verify intermediate results. \rev{In this setting, understanding guides candidate selection rather than directly editing a selected image.} Consequently, the generative capacity remains restricted to outcomes accessible through repeated sampling. In contrast, prompt-based methods, \eg, T2I-R1~\citep{jiang2025t2i}, ReasonGen-R1~\citep{zhang2025reasongen}, and ImageGenCoT~\citep{liao2025imagegen}, leverage CoT to refine the input prompt by decomposing it into semantic aspects, to further improve the prompt for generation. \rev{However, this framework is limited to reasoning before generation, with image-specific correction requiring interaction with the generated output.}

\rev{\paragraph{Self-correction for image generation.} Recent studies have explored iterative self-correction for image generation. SLD~\citep{wu2024sld} combines an LLM controller with an open-vocabulary detector and corrects object layouts through latent-space operations. Idea2Img~\citep{yang2024idea2img} uses GPT-4V as an external critic and prompt designer, selecting candidates and revising prompts for a black-box generator. GenArtist~\citep{wang2024genartist} uses an MLLM agent to plan, verify, and correct results through a library of generation and editing tools. Reflect-DiT~\citep{li2025reflectdit} trains a context-conditioning mechanism that incorporates previous images and textual feedback into a diffusion transformer. 
\citet{zhuo2025reflectionflow} introduce ReflectionFlow, which uses reflection tuning to train a diffusion-based image corrector.
}

\rev{
In contrast to these works, T-GUI uses a single UMM to perform generation, verification, and image editing, without additional training or reliance on an external critic, detector, or tool library.
In addition, compared with Idea2Img, which revises prompts based on feedback to regenerate candidate images, T-GUI utilizes feedback as an instruction to edit the preceding image. In Section~\ref{ablation}, we compare these two \textit{``understanding-generation”} interaction frameworks under a matched feedback-generation procedure. 
Our proposed T-GUI demonstrates how \textit{``understanding-generation”} interaction supports self-correction within an existing UMM.  
}


\section{Methodology}

\subsection{Problem Formulation}
Text-to-image generation aims to synthesize the image $I$ from the given text prompt $t$ by the generative model $f$:
\begin{equation}
    I \ = \ f( t ) \ .
\end{equation}
However, the generative models struggle with complex prompts requiring compositional understanding and spatial reasoning. To address this, the reasoning methods are proposed for text-to-image generation to decompose the generation process into sequential augmented steps:
\begin{equation}
    h_{i} \ = \
\begin{cases} 
   G \ ( \ t,\phi \ ), &  \ \ i\in \{1\} \\
   G \ ( \ t, h_{i-1} \ ), & \ \ i \in \{2,3, ..., n\},
\end{cases}
\end{equation}
where $G(\cdot)$ is the augmented framework for each step, and $h_i$ is the intermediate stage of step $i$.
Finally, the final image $I$ is then conditioned on the complete reasoning chain:
\begin{equation}
    I \ = \ f( \ t \ | \ \{ \ h_{1}, h_{2}, ..., h_{n} \ \} \ ) \ .
\end{equation}


\subsection{Test-time Generation-Understanding Interaction}

\begin{figure*}[t!]
    \centering
    \includegraphics[width=\textwidth]{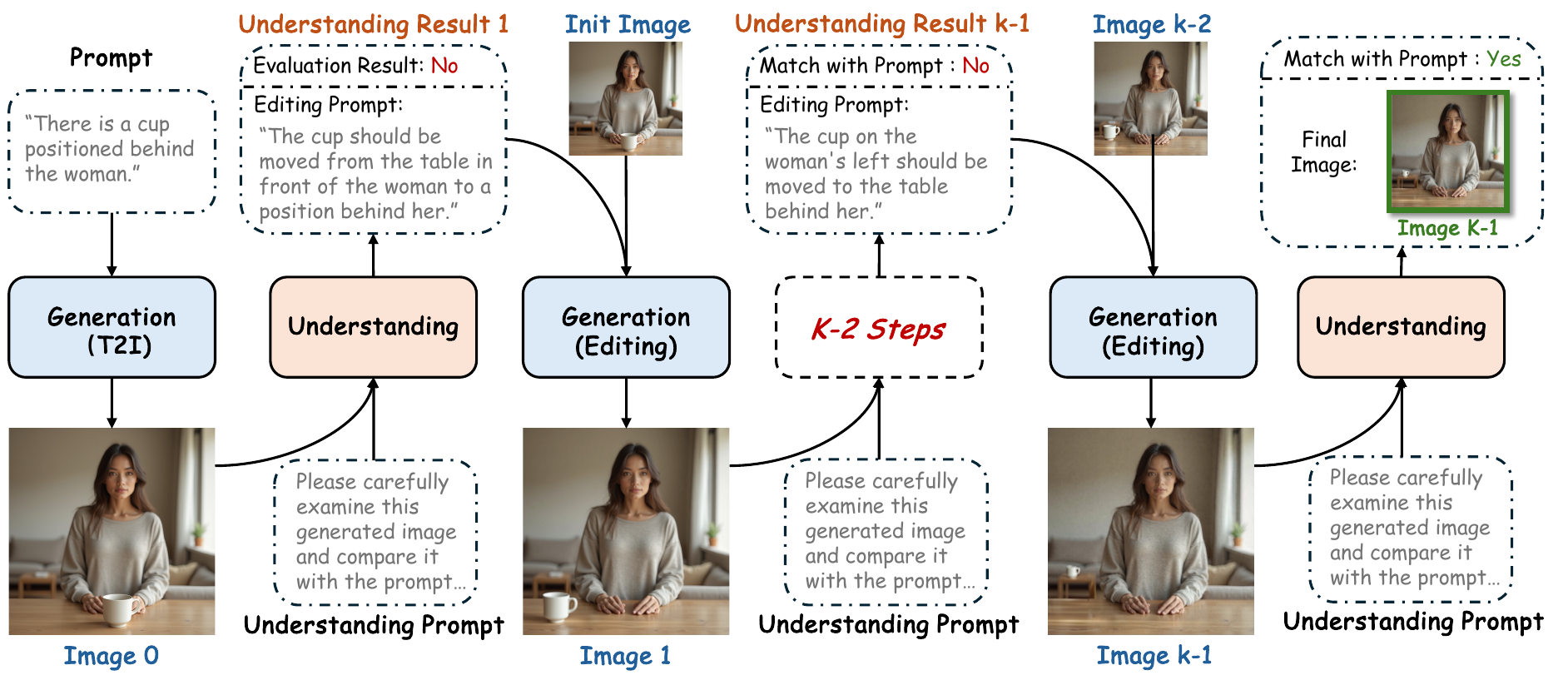}
    \caption{The overall framework of our proposed T-GUI framework. The core insight is to iteratively incorporate understanding-driven guidance and self-verification into the generation process, enabling the UMM to identify and correct generation at test time.}
    \label{fig: framework}
\end{figure*}

We present Test-time Generation-Understanding Interaction in Figure~\ref{fig: framework}. T-GUI enhances the generative capabilities of unified models by infusing their strong understanding abilities into the step-by-step generative process.
The core idea of our T-GUI is to iteratively incorporate understanding-driven guidance and self-verification into the generation process.
\rev{Throughout this paper, one iteration means one executed edit. The initial image $Image_0$ is not an iteration; $K$ denotes the maximum number of edits, and $K=0$ recovers the base generator.}
Specifically, given an input text prompt \( Prompt \), we first generate an initial image \( Image_{0} \) using the original generative function $\text{Generate}_{t2i}(\cdot)$ of the UMM.

Next, we combine the initial image \( Image_{0} \) with an understanding prompt \( Prompt^{\text{un}} \) (\textit{see Appendix for the full prompt}) as input to the understanding module. This gives an evaluation result \( R^{\text{match}}_{1} \) and an editing prompt \( Prompt^{\text{edit}}_{1} \), formulated as:
\begin{equation}
    R^{\text{match}}_{i}, \ Prompt^{\text{edit}}_{i} = \text{Und} \left( \left\{ Image_{i-1}, \ Prompt^{\text{un}} \right\} \right),
\end{equation}
where \( \text{Und}(\cdot) \) denotes the understanding function of the UMM, and \( i \) is the step index (with \( i = 1 \) in this case).

\rev{At step $i$, the understanding module judges whether $Image_{i-1}$ matches the prompt. A \textit{``Yes”} judgment terminates refinement, while a \textit{``No”} judgment triggers an edit.}

As illustrated in Figure~\ref{fig: framework}, we incorporate the strong understanding capability into the generation process via the following formulation:
\begin{equation}
    Image_{i} = \text{Gen}_{editing}\left( \left\{ Image_{i-1}, \ Prompt^{\text{edit}}_{i} \right\} \right),
\end{equation}

\rev{where $\text{Gen}_{editing}(\cdot)$ denotes the image-editing function of the same UMM.} 
\rev{Image editing applies the understanding-derived instruction while conditioning on the previous image. Its effectiveness depends on the reliability of the diagnosis and the preservation of satisfied requirements.}

\rev{Let $\mathcal{A}=\{k\in\{1,\ldots,K\}:R_k^{\mathrm{match}}=\mathrm{Yes}\}$ contain the accepted verification steps, and let $\tau=\min\mathcal{A}$ when this set is nonempty. The final output is}
{\color{black}
\begin{equation}
 Image_{\mathrm{final}}=\begin{cases}
 Image_{\tau-1}, & \mathcal{A}\ne\varnothing,\\
 Image_K, & \mathcal{A}=\varnothing.
 \end{cases}
 \label{eq:final_output}
\end{equation}
}
\rev{At step $k$, the model verifies $Image_{k-1}$ and returns it if accepted. Otherwise, it performs the $k$-th edit to produce $Image_k$. Once the budget of $K$ edits is exhausted, the model returns $Image_K$.}



\begin{figure}[t!]
  \centering
  \includegraphics[width=0.7\linewidth]{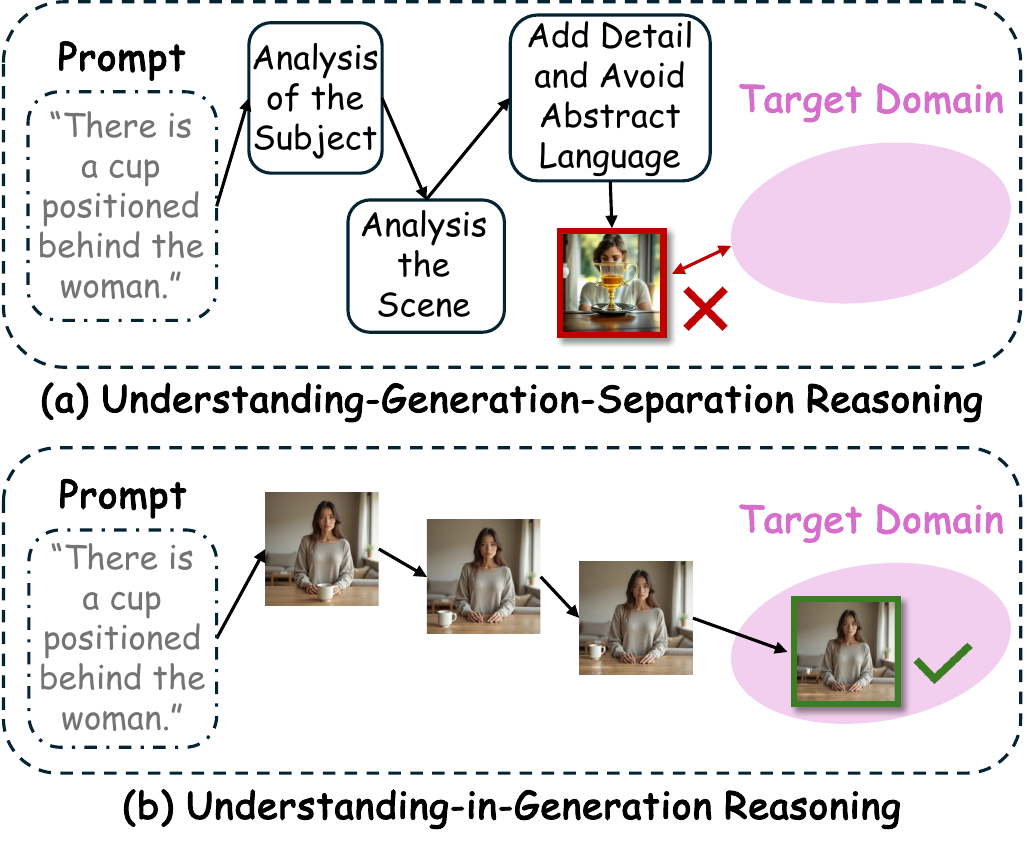}
  \caption{The comparison of reasoning processes between the \textbf{(a)} Understanding-Generation-Separation and \textbf{(b)} Understanding-in-Generation reasoning.}
  \label{fig: why_uigen}
\end{figure}

\subsection{Why We Need Generate-Understand Interaction?}


We explore the intuition behind our proposed T-GUI by addressing a central question: ``\textit{Why is understanding helpful in generation?}"
As demonstrated in Figure~\ref{fig: why_uigen} \textbf{(a)}, the Generation-Understanding separation approach utilizes a Chain-of-Thought strategy to decompose the prompt construction process into discrete components, \eg, subject, scene requirements, and details. This approach follows a ``\textit{first understand, then generate}" paradigm, in which understanding and generation are explicitly decoupled.
However, this separation restricts the reasoning process from incorporating the generative limitations of the base model. 
Consequently, as depicted in Figure~\ref{fig: why_uigen} \textbf{(a)}, this framework fails to guide prompt refinement in ways that address the generative weakness of the base model, \eg, the incorrect spatial relationship between the cup and the woman.
In contrast, our Generation-Understanding interaction framework leverages the strong understanding capability of the base model to directly guide and correct the generative direction during inference. As illustrated in Figure~\ref{fig: why_uigen} \textbf{(b)}, the understanding of the model identifies problems in the initial output, \eg, the incorrect relative spatial position between the cup and the woman, and guides the generative direction to overcome these problems. 

\subsection{Implementation}

\noindent \textbf{Base Model:}
\rev{We adopt the public BAGEL-7B-MoT~\citep{deng2025emerging} as the default base model for generation, understanding, and editing without fine-tuning. Additional comparisons using other UMM backbones are provided in Appendix~\ref{Appendix: Additional_visual_comparison}. BAGEL uses a Mixture-of-Transformers architecture for autoregressive text generation and diffusion-based image generation. 
Full inference settings are listed in Appendix~\ref{app:inference_settings}.}

\noindent \textbf{Understanding Prompt:}
We have designed an understanding prompt to enhance the cognitive and analytical capabilities of UMM. In this way, the base model identifies the weaknesses in the generated images and specifies guidance for improvement in the visual editing. The full text of the prompt is provided \textit{in Section~\ref{Appendix: Details_of_UiG} of the Appendix}.

\noindent \textbf{Iteration:}
\rev{We set the default maximum number of edits to $K=4$. As shown in Table~\ref{tab: ablation_iter}, $K=4$ achieves the highest overall score among the tested budgets, although individual refinement trajectories may still deteriorate. BAGEL without editing corresponds to $K=0$.}

\begin{table*}[t!]
\renewcommand{\tabcolsep}{2pt}
\caption{\textcolor{black}{Evaluation results on TIIF-Bench. Attr., Rela., and Reas. denote Attribute, Relation, and Reasoning. 
Both BAGEL and T-GUI use an inference mode without CoT reasoning.
}}
\resizebox{\linewidth}{!}{
\begin{tabular}{l|r|rrrr|rrrrrr|r}
\toprule
\multirow{3}{*}{\textbf{Model}} & \multirow{3}{*}{\textbf{Overall}}& \multicolumn{4}{c|}{\textbf{Basic Following}} & \multicolumn{6}{c|}{\textbf{Advanced Following}} & \textbf{Designer} \\
\cmidrule{3-13}
& & \multirow{2}{*}{\textbf{Avg.}} & \multirow{2}{*}{\textbf{Attr.}} & \multirow{2}{*}{\textbf{Rela.}} & \multirow{2}{*}{\textbf{Reas.}} & \multirow{2}{*}{\textbf{Avg.}} & \textbf{Attr.} & \textbf{Attr.} & \textbf{Rela.} & \multirow{2}{*}{\textbf{Style}} & \multirow{2}{*}{\textbf{Text}} & \textbf{Real} \\
& & & & & & & \textbf{+Rela.} & \textbf{+Reas.} & \textbf{+Reas.} & & & \textbf{World} \\
\midrule

\multicolumn{13}{c}{\textbf{Short Prompt Setting}} \\
\midrule
Llamagen    & 41.67 & 53.00 & 48.33 & 59.57 & 51.07 & 35.89 & 38.82 & 40.84 & 49.59 & 46.67 & 0.00 & 39.73 \\
LightGen    & 53.22 & 66.58 & 55.83 & 74.82 & 69.07 & 46.74 & 62.44 & 61.71 & 50.34 & 53.33 & 0.00 & 50.92 \\
Show-o      & 59.72 & 73.08 & 74.83 & 78.82 & 65.57 & 53.67 & 60.95 & 68.59 & 66.46 & 63.33 & 3.83 & 55.02 \\
Infinity    & 62.07 & 73.08 & 74.33 & 72.82 & 72.07 & 56.64 & 60.44 & \textbf{74.22} & 60.22 & 80.00 & 10.83 & 54.28 \\
Janus-Pro   & 66.50 & 79.33 & 79.33 & 78.32 & \textbf{80.32} & 59.71 & 66.07 & 70.46 & 67.22 & 60.00 & 28.83 & 65.84 \\
Bagel       & 68.25 & 74.84 & 76.50 & 79.00 & 69.00 & 64.58 & 65.83 & 61.88 & 66.41 & \textbf{90.00} & \textbf{36.20} & 69.40 \\
\midrule
ImageCoT    & 47.41 & 61.60 & 63.00 & 63.57 & 58.23 & 48.94 & 60.95 & 46.53 & 46.36 & 40.00 & 1.81 & 46.27 \\
ReasonGen   & 61.66 & 76.55 & 79.50 & 74.49 & 75.65 & 62.36 & 66.43 & 72.28 & 57.27 & 40.00 & 13.57 & 75.75 \\
T2I-R1      & 68.59 & \textbf{82.90} & \textbf{86.50} & \textbf{83.47} & 78.73 & \textbf{69.05} & 71.64 & 72.43 & \textbf{69.40} & 60.00 & 27.60 & 67.54 \\
\rowcolor{gray!20} \textbf{Our T-GUI}     & \textbf{69.70} & 80.40 & 84.50 & 80.58 & 76.13 & 67.74 & \textbf{73.84} & 66.59 & 65.95 & 80.00 & 30.32 & \textbf{69.40} \\
\midrule
\multicolumn{13}{c}{\textbf{Long Prompt Setting}} \\
\midrule
Llamagen    & 38.22 & 50.00 & 42.33 & 60.32 & 47.32 & 32.61 & 31.57 & 47.22 & 46.22 & 33.33 & 0.00 & 35.62 \\
LightGen    & 43.41 & 47.91 & 47.33 & 45.82 & 50.57 & 41.53 & 40.82 & 50.47 & 45.34 & 53.33 & 6.83 & 50.55 \\
Show-o      & 58.86 & 75.83 & 79.83 & 78.32 & 69.32 & 50.38 & 56.82 & 68.96 & 56.22 & 66.67 & 2.83 & 50.92 \\
Infinity    & 62.32 & 75.41 & 76.83 & 77.57 & 71.82 & 54.98 & 55.57 & 64.71 & 59.71 & 73.33 & 23.83 & 56.89 \\
Janus-Pro   & 65.02 & 78.25 & 82.33 & 73.32 & 79.07 & 58.82 & 56.20 & 70.84 & 59.97 & 70.00 & 33.83 & 60.25 \\
Bagel       & 66.54 & 77.09 & \textbf{83.00} & 74.54 & 72.72 & 64.67 & 67.90 & 64.01 & 65.81 & 66.67 & 33.03 & 70.15 \\
\midrule
ImageCoT    & 45.41 & 59.23 & 59.50 & 56.97 & 61.23 & 44.76 & 53.80 & 41.14 & 45.15 & 40.00 & 3.17 & 47.76 \\
ReasonGen   & 65.11 & 77.14 & 81.00 & 75.72 & 74.69 & 65.83 & 70.81 & 70.99 & 61.96 & 53.33 & 28.05 & 69.40 \\
T2I-R1      & 67.19 & \textbf{81.63} & \textbf{83.00} & \textbf{79.43} & \textbf{82.46} & 68.00 & 69.47 & \textbf{69.95} & 70.40 & 63.33 & 26.24 & 60.45 \\
\rowcolor{gray!20} \textbf{Our T-GUI}     & \textbf{71.11} & 79.05 & \textbf{83.00} & 75.83 & 78.33 & \textbf{70.36} & \textbf{73.36} & 67.59 & \textbf{73.27} & \textbf{76.67} & \textbf{36.65} & \textbf{75.00} \\
\bottomrule
\end{tabular}
}
\label{tab: tiif_results}
\vspace{-12pt}
\end{table*}

\begin{table*}[t!]
\centering
\caption{\rev{Evaluation results on the WISE benchmark.}}
\resizebox{\linewidth}{!}{
\begin{tabular}{lrrrrrrr}
\toprule
Model & Overall & Cultural & Time & Space & Biology & Physics & Chemistry \\
\midrule
Janus-1.3B & 0.23 & 0.16 & 0.26 & 0.35 & 0.28 & 0.30 & 0.14 \\
Janus-Pro-1B & 0.26 & 0.20 & 0.28 & 0.45 & 0.24 & 0.32 & 0.16 \\
Orthus-7B-instruct & 0.27 & 0.23 & 0.31 & 0.38 & 0.28 & 0.31 & 0.20 \\
vila-u-7B & 0.31 & 0.26 & 0.33 & 0.37 & 0.35 & 0.39 & 0.23 \\
Show-o & 0.35 & 0.28 & 0.40 & 0.48 & 0.30 & 0.46 & 0.30 \\
Janus-Pro-7B & 0.35 & 0.30 & 0.37 & 0.49 & 0.36 & 0.42 & 0.26 \\
Emu3 & 0.39 & 0.34 & 0.45 & 0.48 & 0.41 & 0.45 & 0.27 \\
T2I-R1 & 0.54 & 0.56 & 0.55 & 0.63 & 0.54 & 0.55 & 0.30 \\
\midrule
\rev{BAGEL} & \rev{0.52} & \rev{0.44} & \rev{0.55} & \rev{0.68} & \rev{0.44} & \rev{0.60} & \rev{0.39} \\
\rev{BAGEL w/ CoT} & \rev{0.70} & \rev{0.76} & \rev{0.69} & \rev{0.75} & \rev{0.65} & \rev{0.75} & \rev{0.58} \\
\rev{T-GUI on BAGEL} & \rev{0.72} & \rev{0.77} & \rev{0.69} & \rev{0.75} & \rev{0.66} & \rev{0.76} & \rev{0.61} \\
\rev{T-GUI on BAGEL w/ CoT} & \rev{0.74} & \rev{0.80} & \rev{0.72} & \rev{0.76} & \rev{0.71} & \rev{0.79} & \rev{0.63} \\
\bottomrule
\end{tabular}}
\label{tab: wise_results}
\end{table*}

\section{Experiment}

\subsection{Benchmark and Experiment Setup}

\noindent \textbf{Benchmark:}
We evaluate the performance of text-to-image generation using the TIIF benchmark~\citep{wei2025tiif} and the WISE benchmark~\citep{niu2025wise}. The TIIF benchmark provides a comprehensive framework for fine-grained assessment of text-to-image models. It features 36 novel prompt combinations spanning six compositional dimensions, along with 100 real-world, designer-level prompts that demand sophisticated aesthetic judgment. The WISE benchmark, a widely adopted standard in this domain, poses challenges by embedding world knowledge within prompts, thereby testing a model’s capacity for knowledge-based text-to-image reasoning.

\noindent \textbf{Baselines:}
Our baseline models consist of original AR-based models, including Llamagen~\citep{sun2024autoregressive}, LightGen~\citep{wu2025lightgen}, Show-o~\citep{xieshow}, Infinity~\citep{han2025infinity}, Janus-Pro~\citep{chen2025janus}, Orthus~\citep{kou2024orthus}, Vila-u~\citep{wu2024vila}, and Emu3~\citep{wang2024emu3}; as well as augmented text-to-image models, including ImageCoT~\citep{zhang2025let}, ReasonGen-R1~\citep{zhang2025reasongen}, and T2I-R1~\citep{jiang2025t2i}.

\noindent \textbf{Experiment Setting:}
\rev{We use the original TIIF-Bench and original WISE evaluation protocols. TIIF-Bench baselines are regenerated and evaluated under our protocol; WISE baselines other than BAGEL use published results. BAGEL and T-GUI use the official WISE prompts, GPT-4o judging, and WiScore. Inference and evaluation details are provided in Appendix~\ref{app:inference_settings}.} 
For more details about these two benchmarks, \textit{please refer to Section~\ref{Appendix: Details_of_experiment} in the Appendix}.

\begin{figure*}[t!]
    \centering
    \includegraphics[width=\textwidth]{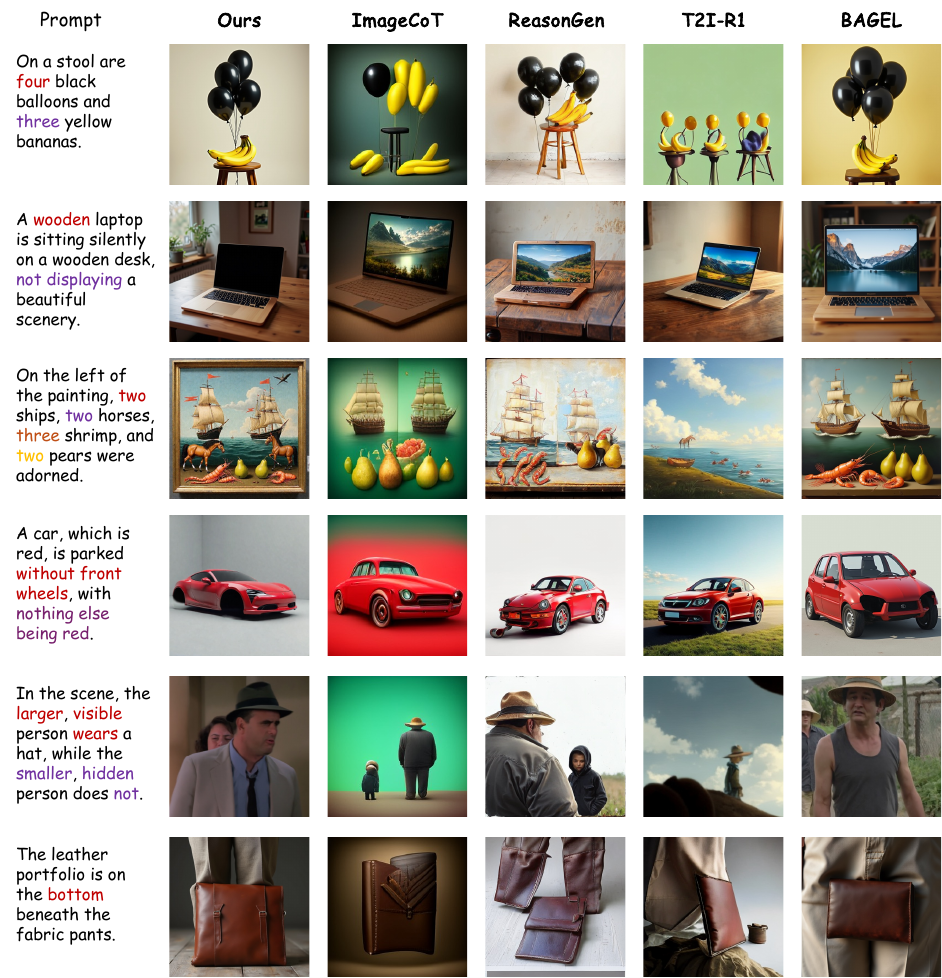}
    \vspace{-14pt}
    \caption{The visual comparison between the existing augmented frameworks and our proposed T-GUI framework.}
    \label{fig: short_result}
    \vspace{-12pt}
\end{figure*}

\subsection{Quantitative Results}

\rev{As shown in Table~\ref{tab: tiif_results}, compared with the BAGEL backbone, our T-GUI improves the overall score by \textbf{1.45 points} under short prompts and \textbf{4.57 points} under long prompts. 
}

\rev{We report the WISE results of BAGEL and T-GUI with and without native CoT in Table~\ref{tab: wise_results}. The results show that T-GUI improves BAGEL's performance both without and with native CoT, increasing the overall scores from 0.52 to 0.72 ($+0.20$) and from 0.70 to 0.74 ($+0.04$), respectively. These gains demonstrate the effectiveness of T-GUI under both inference settings.}

To quantify uncertainty in these gains, we report paired 95\% confidence intervals in Table~\ref{tab:paired_ci}. For each of 2,000 bootstrap samples, we draw prompts with replacement within each WISE category or TIIF-Bench prompt type, keeping the original number of prompts in each group. We use the recorded scores for the same sampled prompts to recompute the difference in overall score between T-GUI and BAGEL. The interval bounds are the 2.5th and 97.5th percentiles of these differences.

\begin{table}[!htbp]
\centering
\vspace{-6pt}
\caption{Overall score gains over BAGEL at the default refinement budget ($K=4$).}
\label{tab:paired_ci}
\small
\begin{tabular*}{\linewidth}{@{\extracolsep{\fill}}l|l|rrrr@{}}
\toprule
Benchmark & Setting & BAGEL & T-GUI & Gain & 95\% CI for gain \\
\midrule
WISE & CoT off & 0.52 & 0.72 & $+0.20$ & $[+0.17, +0.24]$ \\
WISE & CoT on & 0.70 & 0.74 & $+0.04$ & $[+0.01, +0.06]$ \\
TIIF-Bench & Short prompts & 68.25 & 69.70 & $+1.45$ & $[+0.32, +2.56]$ \\
TIIF-Bench & Long prompts & 66.54 & 71.11 & $+4.57$ & $[+3.16, +5.84]$ \\
\bottomrule
\end{tabular*}
\end{table}

At the default editing budget ($K=4$), all four intervals lie above zero. On WISE with native CoT, the interval ranges from $+0.01$ to $+0.06$, indicating a small average gain. As shown in Section~\ref{ablation}, increasing the budget from $K=4$ to $K=5$ lowers the overall score for both short and long prompts on TIIF-Bench.

\rev{Finally, we analyze the trade-off between the performance gains achieved by applying T-GUI to the base model and the associated inference latency. Detailed results are reported in Table~\ref{tab: trade-off} in Appendix~\ref{app:trade_off}.}

\begin{figure*}[t!]
    \centering
    \begin{overpic}[width=\textwidth]{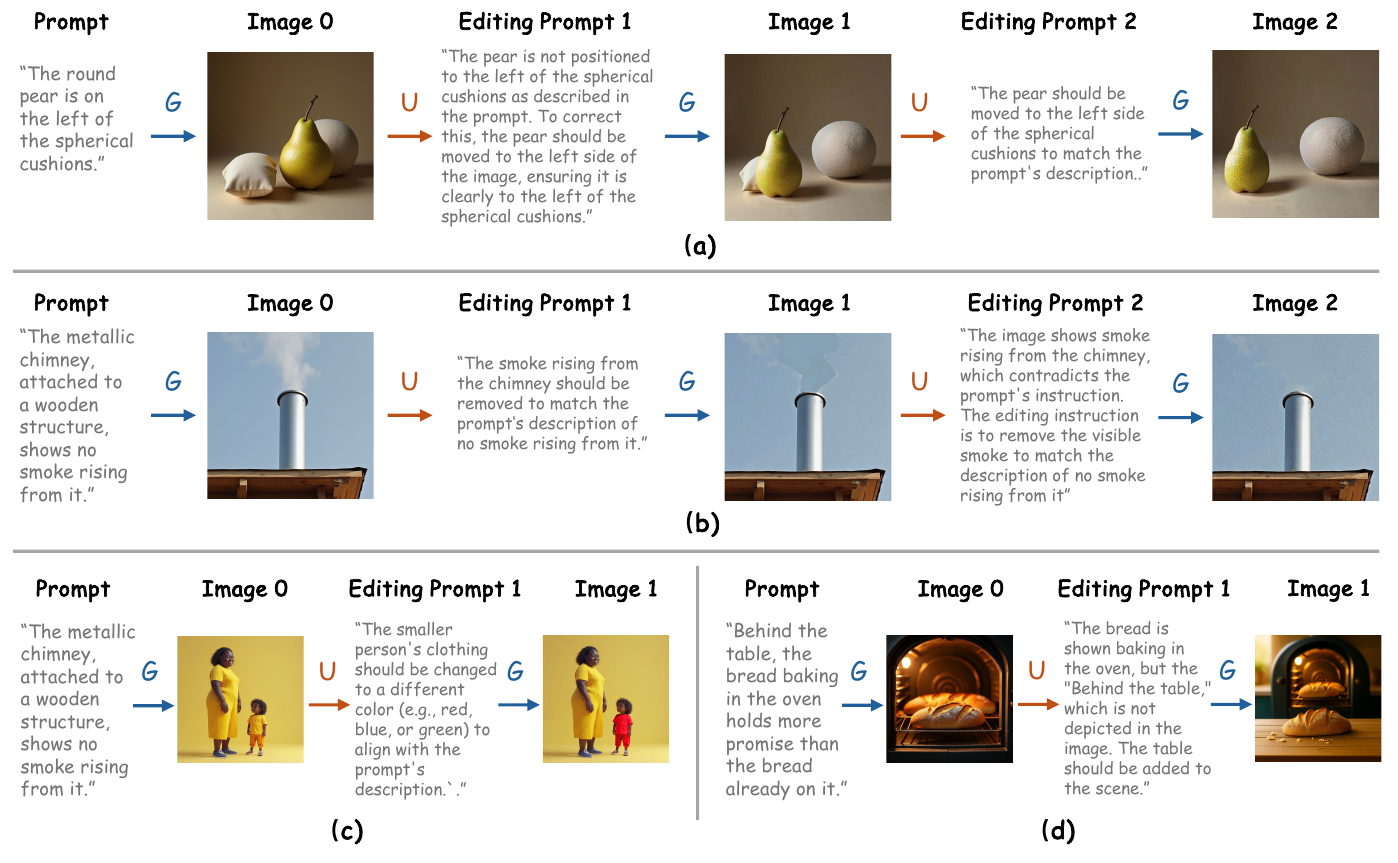}
      \put(1.2,3.8){\color{white}\rule{0.089\textwidth}{0.140\textwidth}}
      \put(1.3,17.2){\makebox(0,0)[lt]{\parbox[t]{0.087\textwidth}{\color{black}\raggedright\sffamily\fontsize{6}{7}\selectfont A larger person wearing yellow clothing stands next to a smaller person dressed in a different color.}}}
    \end{overpic}
    \vspace{-20pt}
    \caption{The visualization of the self-correction within our T-GUI framework. \rev{\textbf{(a)} and \textbf{(b)} show two-edit cases, while \textbf{(c)} and \textbf{(d)} show one-edit cases. The initial image $Image_0$ is excluded from the iteration count.} \textit{Additional visualizations are provided in Section~\ref{Appendix: Additional_reasoning_visualization} of the Appendix}.}
    \label{fig: reasoning}
    \vspace{-4pt}
\end{figure*}

\subsection{Qualitative Results}



As illustrated in Figure~\ref{fig: short_result}, we present a visual comparison between existing augmented text-to-image generation methods (\eg, ImageCoT~\citep{zhang2025let}, ReasonGen-R1~\citep{zhang2025reasongen}, and T2I-R1~\citep{jiang2025t2i}) and our proposed T-GUI framework.
The visual results demonstrate a notable improvement in prompt alignment achieved by T-GUI in the generated images. For instance, given the prompt {\fontfamily{qcr}\selectfont["On a stool are four black balloons and three yellow bananas"]}, only T-GUI accurately generates the correct quantities for both balloons and bananas, as shown in Figure~\ref{fig: short_result}.
Furthermore, in response to the prompt {\fontfamily{qcr}\selectfont["A wooden laptop is sitting silently on a wooden desk, not displaying a beautiful scenery"]}, all existing methods fail to generate an image without displaying scenery. In contrast, T-GUI produces a visually accurate result consistent with the prompt, as also shown in Figure~\ref{fig: short_result}.
These visual comparisons show the substantial gains in prompt following and generation accuracy provided by our T-GUI framework. Additional visual comparison results are presented \textit{in Section~\ref{Appendix: Additional_visual_comparison} of the Appendix}.


Additionally, we present a visualization of the self-correction process within our T-GUI framework, as shown in Figure~\ref{fig: reasoning}. In case \textbf{(a)}, T-GUI effectively identifies a weakness in the original generated image, specifically the incorrect spatial relationship between the pear and the spherical cushions, which is attributed to its strong understanding capability. Subsequently, our T-GUI uses ``\textit{Image Editing}" as a bridge to guide the generative direction for the weakness in spatial relationship, and generates the well-matched image.

\section{Ablation Study}
\label{ablation}
\begin{table*}[t!]
\renewcommand{\tabcolsep}{3pt}
\caption{\textcolor{black}{\textbf{Ablation on understanding-derived feedback.} We report the results on the TIIF benchmark using either the original prompt or understanding-derived feedback as the editing instruction. Both variants retain image editing.}}
\resizebox{\linewidth}{!}{
\begin{tabular}{
>{\centering\arraybackslash}p{1.5cm}| 
>{\centering\arraybackslash}p{1.2cm}| 
>{\centering\arraybackslash}p{1.2cm} 
>{\centering\arraybackslash}p{1.2cm} 
>{\centering\arraybackslash}p{1.2cm} 
>{\centering\arraybackslash}p{1.2cm}| 
>{\centering\arraybackslash}p{1.2cm} 
>{\centering\arraybackslash}p{1.2cm} 
>{\centering\arraybackslash}p{1.2cm} 
>{\centering\arraybackslash}p{1.2cm} 
>{\centering\arraybackslash}p{1.2cm} 
>{\centering\arraybackslash}p{1.2cm}| 
>{\centering\arraybackslash}p{1.3cm} 
}
\toprule
\textcolor{black}{\textbf{Targeted}} & \multirow{3}{*}{\textbf{Overall}} & \multicolumn{4}{c|}{\textbf{Basic Following}} & \multicolumn{6}{c|}{\textbf{Advanced Following}} & \textbf{Designer} \\
\cmidrule{3-13}
\textcolor{black}{\textbf{Feedback}} & & \multirow{2}{*}{\textbf{Avg.}} & \multirow{2}{*}{\textbf{Attr.}} & \multirow{2}{*}{\textbf{Rela.}} & \multirow{2}{*}{\textbf{Reas.}} & \multirow{2}{*}{\textbf{Avg.}} & \textbf{Attr.} & \textbf{Attr.} & \textbf{Rela.} & \multirow{2}{*}{\textbf{Style}} & \multirow{2}{*}{\textbf{Text}} & \textbf{Real} \\
 & & & & & & & \textbf{+Rela.} & \textbf{+Reas.} & \textbf{+Reas.} & & & \textbf{World} \\
\midrule

\multicolumn{13}{c}{\textbf{Short Prompt Setting}} \\
\midrule
\xmark    & 64.30 & 77.70 & 81.50 & 80.05 & 71.56 & 65.18 & 69.75 & 65.06 & \textbf{66.98} & 70.00 & 10.41 & 63.43 \\
\cmark    & \textbf{69.70} & \textbf{80.40} & \textbf{84.50} & \textbf{80.58} & \textbf{76.13} & \textbf{67.74} & \textbf{73.84} & \textbf{66.59} & 65.95 & \textbf{80.00} & \textbf{30.32} & \textbf{69.40}  \\
$\Delta $ & \upx{5.40} & \upx{2.70} & \upx{3.00} & \upx{0.53} & \upx{4.57} & \upx{2.56} & \upx{4.09} & \upx{1.53} & \downx{1.03} & \upx{10.00} & \upx{19.91} & \upx{5.97} \\
\midrule
\multicolumn{13}{c}{\textbf{Long Prompt Setting}} \\
\midrule
\xmark    & 65.04 & 78.14 & 80.50 & \textbf{81.97} & 71.96 & 66.24 & 69.35 & \textbf{68.00}& 68.08 & 63.33 & 15.38 & 66.79 \\
\cmark    & \textbf{71.11} & \textbf{79.05} & \textbf{83.00} & 75.83 & \textbf{78.33} & \textbf{70.36} & \textbf{73.36} & 67.59 & \textbf{73.27} & \textbf{76.67} & \textbf{36.65} & \textbf{75.00}  \\
$\Delta $ & \upx{6.07} & \upx{0.91} & \upx{2.50} & \downx{6.14} & \upx{6.37} & \upx{4.12} & \upx{4.01} & \downx{0.41} & \upx{5.19} & \upx{13.34} & \upx{21.27} & \upx{8.21} \\
\bottomrule
\end{tabular}
}
\label{tab: ablation_edit}
\vspace{-8pt}
\end{table*}

\begin{figure*}[t!]
\centering
\includegraphics[width=\textwidth]{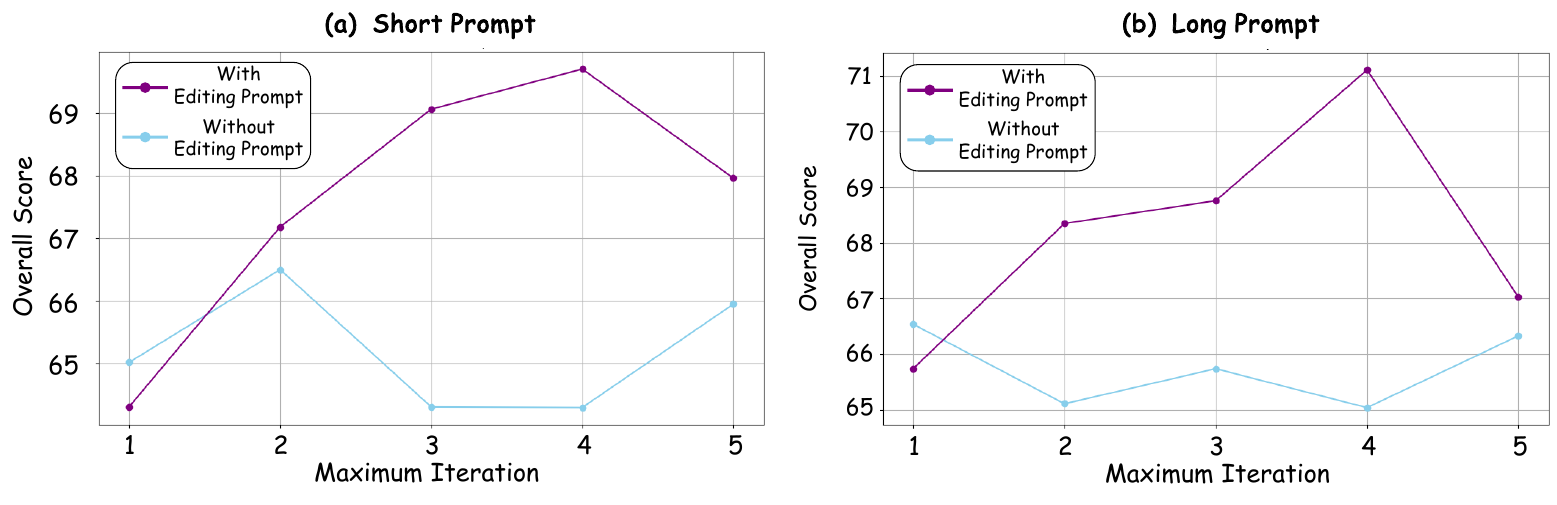}
\caption{\rev{Ablation of feedback across maximum editing budgets $K$. Both variants retain the image-editing operator; the comparison is between understanding-derived editing instructions and the original prompt. Panels (a) and (b) show short and long prompts. The initial image is excluded from the iteration count.}}
\label{fig: trend}
\end{figure*}



\rev{\paragraph{Feedback and correction operator.}
We examine the roles of understanding-derived feedback and the correction operator in T-GUI. Table~\ref{tab: ablation_edit} evaluates the contribution of targeted feedback while retaining image editing in both variants. The variant without targeted feedback uses the original prompt as the editing instruction. Targeted feedback increases the overall score from 64.30 to 69.70 for short prompts and from 65.04 to 71.11 for long prompts. These gains demonstrate the effectiveness of editing instructions.}

\rev{Furthermore, we examine the correction operator by comparing image editing with regeneration from a revised prompt in Table~\ref{tab:correction_operator}. Both variants start from the same initial image and use the same understanding prompt, verifier, stopping rule, seed, and maximum budget $K=4$. They also follow the same feedback-generation procedure, with each variant generating feedback from its own current image. When the understanding module identifies a mismatch, T-GUI edits the current image using the feedback as its instruction. The regeneration variant instead uses the same UMM to revise the original prompt based on the feedback and then generate a new image without conditioning on the preceding image.}

\begin{table}[t]
\color{black}
\centering
\caption{Ablation of the correction operator on TIIF-Bench at $K=4$.}
\label{tab:correction_operator}
\setlength{\tabcolsep}{3pt}
\resizebox{\linewidth}{!}{%
\begin{tabular}{lrrrrrrrrrrrr}
\toprule
Correction operator & Overall & Basic Avg. & Attr. & Rela. & Reas. & Adv. Avg. & Attr.+Rela. & Attr.+Reas. & Rela.+Reas. & Style & Text & Real World \\
\midrule
\multicolumn{13}{c}{\textit{Short prompts}} \\
\midrule
Regenerate & 65.8 & 78.3 & 82.6 & 80.1 & 72.2 & 65.6 & 72.4 & 65.9 & 65.6 & 73.3 & 14.0 & 66.5 \\
Edit (T-GUI) & 69.7 & 80.4 & 84.5 & 80.6 & 76.1 & 67.7 & 73.8 & 66.6 & 66.0 & 80.0 & 30.3 & 69.4 \\
\midrule
\multicolumn{13}{c}{\textit{Long prompts}} \\
\midrule
Regenerate & 68.1 & 77.2 & 81.1 & 75.2 & 75.4 & 69.6 & 71.7 & 67.5 & 71.8 & 72.4 & 32.0 & 65.6 \\
Edit (T-GUI) & 71.1 & 79.0 & 83.0 & 75.8 & 78.3 & 70.4 & 73.4 & 67.6 & 73.3 & 76.7 & 36.7 & 75.0 \\
\bottomrule
\end{tabular}}
\end{table}

\rev{
As shown in Table~\ref{tab:correction_operator}, image editing outperforms regeneration across all reported dimensions, with overall gains of 3.9 and 3.0 points for short and long prompts. These ablations support the effectiveness of targeted feedback and image editing conditioned on the preceding image. The results further indicate that, with a single UMM serving as the base model, editing yields larger performance gains than regeneration.
}


\begin{table*}[t!]
\renewcommand{\tabcolsep}{3pt}
\caption{\textbf{Ablation on the maximum iteration hyperparameter.} We present an ablation study on the maximum iteration hyperparameter using the TIIF benchmark, varying its value from 1 to 5. The results indicate that a setting of \textit{4 iterations} shows the best performance.}
\resizebox{\linewidth}{!}{
\begin{tabular}{
>{\raggedright\arraybackslash}p{1.5cm}| 
>{\centering\arraybackslash}p{1.2cm}| 
>{\centering\arraybackslash}p{1.2cm} 
>{\centering\arraybackslash}p{1.2cm} 
>{\centering\arraybackslash}p{1.2cm} 
>{\centering\arraybackslash}p{1.2cm}| 
>{\centering\arraybackslash}p{1.2cm} 
>{\centering\arraybackslash}p{1.2cm} 
>{\centering\arraybackslash}p{1.2cm} 
>{\centering\arraybackslash}p{1.2cm} 
>{\centering\arraybackslash}p{1.2cm} 
>{\centering\arraybackslash}p{1.2cm}| 
>{\centering\arraybackslash}p{1.3cm} 
}
\toprule
\multirow{2}{*}{\textbf{Maximum}} & \multirow{3}{*}{\textbf{Overall}} & \multicolumn{4}{c|}{\textbf{Basic Following}} & \multicolumn{6}{c|}{\textbf{Advanced Following}} & \textbf{Designer} \\
\cmidrule{3-13}
\multirow{2}{*}{\textbf{Iteration}} & & \multirow{2}{*}{\textbf{Avg.}} & \multirow{2}{*}{\textbf{Attr.}} & \multirow{2}{*}{\textbf{Rela.}} & \multirow{2}{*}{\textbf{Reas.}} & \multirow{2}{*}{\textbf{Avg.}} & \textbf{Attr.} & \textbf{Attr.} & \textbf{Rela.} & \multirow{2}{*}{\textbf{Style}} & \multirow{2}{*}{\textbf{Text}} & \textbf{Real} \\
& & & & & & & \textbf{+Rela.} & \textbf{+Reas.} & \textbf{+Reas.} & & & \textbf{World} \\
\midrule

\multicolumn{13}{c}{\textbf{Short Prompt Setting}} \\
\midrule
Value = 1    & 64.31 & 76.43 & 76.50 & 75.54 & 78.25 & 66.97 & 73.14 & 65.82 & 69.44 & 66.67 & 7.24 & 67.16 \\
Value = 2    & 67.18 & 77.73 & 82.50 & 81.68 & 69.00 & 65.42 & 72.15 & 62.51 & 65.88 & 70.00 & 26.70 & \textbf{74.25} \\
Value = 3    & 69.06 & 79.15 & 84.00 & 80.85 & 72.60 & \textbf{69.85} & \textbf{77.40} & 64.96 & \textbf{72.60} & 73.33 & 23.08 & 72.76 \\
\rowcolor{gray!20} Value = 4    & \textbf{69.70} & 80.40 & \textbf{84.50} & 80.58 & 76.13 & 67.74 & 73.84 & 66.59 & 65.95 & \textbf{80.00} & \textbf{30.32} & 69.40  \\
Value = 5    & 67.96 & \textbf{82.62} & 83.00 & \textbf{82.53} & \textbf{82.33} & 68.12 & 70.30 & \textbf{68.51} & 71.78 & 63.33 & 23.08 & 66.79 \\
\midrule
\multicolumn{13}{c}{\textbf{Long Prompt Setting}} \\
\midrule
Value = 1    & 65.74 & 75.74 & 78.00 & 79.25 & 69.96 & 65.57 & 70.39 & 65.62 & 65.11 & 73.33 & 15.84 & 67.54 \\
Value = 2    & 68.35 & 78.72 & 78.50 & \textbf{82.02} & 75.65 & 67.66 & 72.15 & 62.93 & 70.34 & 73.33 & 25.34 & 72.76 \\
Value = 3    & 68.76 & \textbf{79.86} & \textbf{86.50} & 77.35 & 75.73 & 68.21 & 70.07 & \textbf{70.01} & 68.88 & 70.00 & 29.41 & 70.90 \\
\rowcolor{gray!20} Value = 4    & \textbf{71.11} & 79.05 & 83.00 & 75.83 & \textbf{78.33} & \textbf{70.36} & \textbf{73.36} & 67.59 & \textbf{73.27} & \textbf{76.67} & \textbf{36.65} & \textbf{75.00}  \\
Value = 5    & 67.03 & 77.34 & 81.00 & 76.28 & 74.73 & 66.80 & 69.13 & 66.28 & 70.39 & 63.33 & 27.15 & \textbf{75.00} \\
\bottomrule
\end{tabular}
}
\label{tab: ablation_iter}
\vspace{-12pt}
\end{table*}

\rev{\paragraph{Iteration budget and non-monotonic improvement.}
In Table~\ref{tab: ablation_iter}, each iteration corresponds to one executed edit, and $K=0$ denotes the initial BAGEL image. As the edit budget increases from $K=1$ to $K=4$, scores rise from 64.31 to 69.70 for short prompts and from 65.74 to 71.11 for long prompts. At $K=5$, the scores drop to 67.96 and 67.03, corresponding to declines of 1.74 and 4.08 points, respectively. These results show that additional refinement does not guarantee monotonic improvement. Individual trajectories may also exhibit degradation, even under the budget that yields the highest aggregate score.
}


\rev{\paragraph{Analysis of degrading edits.}
To analyze failure patterns, we randomly select 30 editing transitions from both prompt settings. In each transition, the updated image receives a lower evaluation score than its predecessor. Ten annotators independently review all 30 transitions. They examine the original prompt, the images before and after editing, the editing instruction, and the full refinement trajectory. Each annotator assigns one or more labels from Table~\ref{tab:degradation_patterns}. This yields 300 annotator--transition assessments, including 280 with one label and 20 with two. No assessment receives three or more labels. In total, 320 labels are assigned. Category percentages are calculated over the 300 assessments and therefore sum to 106.67\%.}

\begin{table}[t]
\color{black}
\centering
\caption{Manual characterization of 30 degrading transitions by 10 annotators.}
\label{tab:degradation_patterns}
\begin{tabular}{p{0.67\linewidth}rr}
\toprule
Observed pattern & Assessments & Percentage \\
\midrule
Incorrect feedback requests changes to content that already satisfies the prompt & 51 & 17.00\% \\
Feedback identifies a genuine error, but the requested correction is not successfully executed & 125 & 41.67\% \\
An edit corrects the targeted issue while damaging another previously satisfied requirement & 95 & 31.67\% \\
Successive edits reverse earlier changes or oscillate between competing requirements & 49 & 16.33\% \\
\bottomrule
\end{tabular}
\end{table}

\rev{Unsuccessful execution of a valid correction is the most frequently assigned label (41.67\%), followed by damage to previously correct content (31.67\%). Incorrect feedback and reversal or oscillation receive 17.00\% and 16.33\%, respectively. These observations highlight limitations in feedback reliability and editing accuracy, including the preservation of satisfied requirements. They characterize selected degrading transitions, rather than estimating failure rates over all edits.}


\rev{\paragraph{Selecting an intermediate output.}
To mitigate the observed degradation, we evaluate an intermediate-output selection strategy. We retain the initial image and the subsequent images along each refinement trajectory, and use the same UMM to select the candidate judged to best satisfy the original prompt. We use the same generation trajectories as the original T-GUI and modify only the final output selection. The benchmark evaluator is used solely to evaluate the outputs selected by the UMM.}

\begin{table}[t]
\color{black}
\centering
\caption{Effect of intermediate-output selection on overall TIIF-Bench scores. Original T-GUI follows Eq.~\ref{eq:final_output}; the selection variant chooses among stored trajectory outputs using the same UMM.}
\label{tab:intermediate_selection}
\begin{tabular}{lrrr}
\toprule
Prompt setting & Maximum budget $K$ & Original T-GUI & With selection \\
\midrule
Short & 4 & 69.70 & 70.05 \\
Short & 5 & 67.96 & 69.50 \\
Long & 4 & 71.11 & 71.11 \\
Long & 5 & 67.03 & 71.05 \\
\bottomrule
\end{tabular}
\end{table}

\rev{As shown in Table~\ref{tab:intermediate_selection}, the selection strategy improves short-prompt scores by 0.35 and 1.54 points at $K=4$ and $K=5$, respectively. For long prompts, it leaves the $K=4$ score unchanged and improves $K=5$ by 4.02 points. Consequently, the $K=4$ to $K=5$ decline shrinks from 1.74 to 0.55 points for short prompts and from 4.08 to 0.06 points for long prompts. We use a budget of $K$ additional understanding calls to evaluate the stored candidates, without additional image generation. The strategy remains dependent on the model's judgments, and a decline remains under independent evaluation in both prompt settings. We retain the original stopping-rule output as the primary T-GUI result and report selection separately as a mitigation.}

\section{Conclusion}

In this paper, we propose Test-time Generation-Understanding Interaction (T-GUI), an effective self-correction framework for Unified Multimodal Model in text-to-image generation, which uses feedback from the UMM to guide image editing.
\rev{Our T-GUI treats ``\textit{Image Editing}” as the bridge to incorporate understanding into generation, using the same UMM to generate, verify, and edit images without additional training.}
Our T-GUI demonstrates the promising potential of the UMM to enhance the generative capabilities with the guidance from the strong understanding capabilities during inference.
\rev{The experiment results show overall TIIF-Bench gains of 1.45\% and 4.57\% over BAGEL under short and long prompts, and WISE gains of 0.20 without native thinking and 0.04 when combined with native thinking. Controlled ablations support the roles of targeted feedback and editing the previous image.}

\noindent \textbf{Limitation and Future Work:}
\rev{The loop relies on accurate feedback, reliable editing, and preservation of already satisfied requirements. Verification uses the same UMM as generation and remains limited by its understanding capabilities. Budget exhaustion returns an unverified image. Additional iterations can introduce errors or undo earlier gains. Selecting among intermediate candidates reduces observed degradation but still relies on the model's judgments. Our manual analysis covers a small sample of degrading edits and does not estimate overall failure rates. Variability across repeated runs and broader generalization require further study.}
Currently, T-GUI does not support video generative models. Future work will focus on enhancing the reasoning capabilities for video generation.

\rev{\noindent\textbf{Broader impact.} Iterative editing can help users produce images that better reflect their intentions. However, it can also create misleading images by modifying photographs to match false descriptions. Generation and verification use the same UMM, which may fail to detect its own generation errors. The spatial failures in Appendix~\ref{app:failure_cases} illustrate this limitation. In our sample of degrading transitions, annotators identified incorrect feedback in 17.00\% of the 300 individual assessments.}

\clearpage
{
    \small
    \bibliographystyle{tmlr}
    \bibliography{main}
}


\clearpage
\newpage
\appendix
\section{More Details of Test-time Generation-Understanding Interaction}
\label{Appendix: Details_of_UiG}
We show the full understanding prompt in Figure~\ref{fig: under_prompt}.
To evaluate whether a generated image aligns with its text prompt, the understanding prompt first guides the base model in parsing the semantic structure of the prompt, identifying key elements such as the main subject, setting, visual attributes (\eg, clothing, colors, lighting), and emotional tone. Then the base model should be prompted to map these components directly onto the visual elements present in the image, enabling a structured cross-modal comparison. By explicitly attending to mismatches between the described and rendered attributes, the base model recognizes omissions (\eg, a missing object, a setting, or a background), inconsistencies (\eg, incorrect spatial relationship or incorrect color for some objects), or misrepresentations of mood and style. These observations should then be synthesized into a coherent editing prompt that specifies the required visual adjustments. This process ensures that the editing prompt is grounded in a detailed diagnostic analysis of the weakness of the generated image, thus facilitating well-matched image refinement in subsequent reasoning processes.


\section{More Details of Experiment}
\label{Appendix: Details_of_experiment}

\subsection{More Details of Benchmarks}

\noindent \textbf{TIIF Benchmark} is a comprehensive and fine-grained evaluation benchmark specifically designed to assess the instruction-following capabilities of modern Text-to-Image (T2I) models. Unlike prior benchmarks such as COMPBENCH++~\citep{huang2025t2i} or GENAI BENCH~\citep{li2024genai}, which suffer from semantic redundancy, fixed prompt lengths, and coarse evaluation metrics, TIIF-Bench offers a diverse and hierarchically structured set of 5000 prompts that span a broad range of compositional and semantic complexity.

Each prompt in TIIF-Bench is categorized into three difficulty levels—Basic, Advanced, and Designer-Level Following—and is available in both short and long versions to test prompt-length sensitivity. The prompts are systematically constructed through a two-stage pipeline: (i) concept pool extraction across ten dimensions (attributes, relations, and reasoning), and (ii) attribute composition across 36 defined combinations. In addition to traditional evaluation axes (e.g., color, texture, spatial relations, numeracy), TIIF-Bench introduces three novel dimensions: text rendering, style control, and real-world designer prompts, which reflect practical demands and aesthetic nuance.

TIIF-Bench also introduces a fine-grained, attribute-specific evaluation protocol using large vision-language models (VLMs) like GPT-4o and Qwen-VL2.5. This protocol poses structured yes/no queries to assess semantic alignment without relying on full prompt inclusion, thereby mitigating hallucination effects. For text rendering evaluation, TIIF-Bench proposes a novel metric called Global Normalized Edit Distance (GNED), which robustly measures typographic accuracy by penalizing both over-generation and omission.

Extensive benchmarking across T2I models demonstrates that TIIF-Bench provides deeper diagnostic insight into model robustness, semantic comprehension, and generation fidelity, making it a valuable tool for guiding the development and evaluation of next-generation T2I systems.

\rev{\noindent\textbf{WISE Benchmark.} WISE~\citep{niu2025wise} evaluates world-knowledge-informed text-to-image generation, testing whether a model can infer the visual content implied by a prompt. The original benchmark contains 1,000 prompts spanning 25 subdomains, grouped into Cultural, Time, Space, Biology, Physics, and Chemistry. Its evaluation considers consistency with the intended content, realism, and aesthetic quality using GPT-4o and WiScore.}

\begin{figure}[t!]
    \centering
    \includegraphics[width=0.9\textwidth]{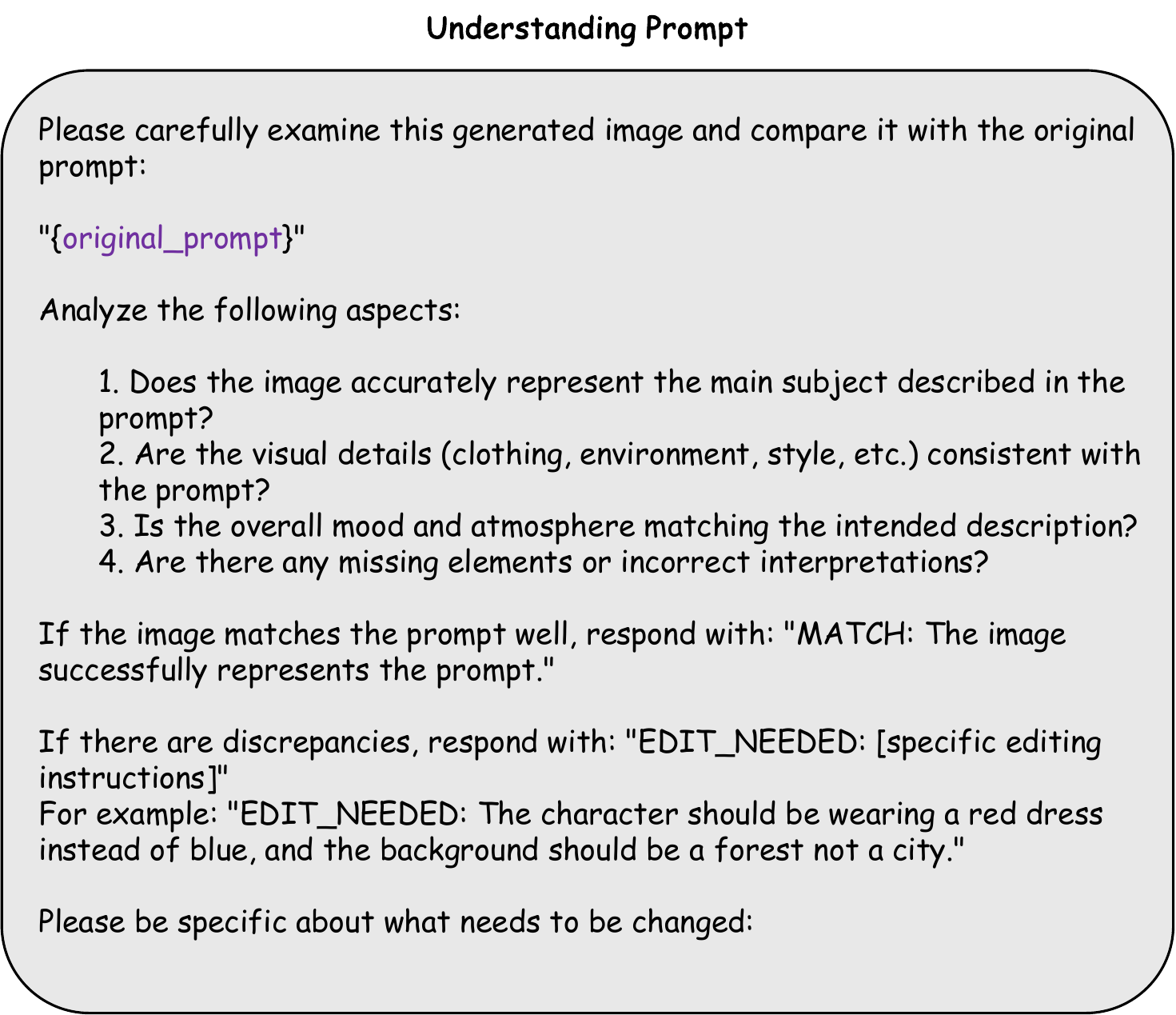}
    \caption{The full understanding prompt}
    \label{fig: under_prompt}
\end{figure}

\subsection{More Details of Experiment Setup}
All reported experiments in this paper were conducted on NVIDIA A100 GPUs. To ensure fair comparisons, we set the random seed to a fixed value of 42. For further code details, please refer to the code provided in \textit{Supplementary Materials}.

\subsection[Inference and Evaluation Settings]{\rev{Inference and Evaluation Settings}}
\label{app:inference_settings}
\rev{\paragraph{Checkpoint and precision.} The following settings apply to experiments using BAGEL. We use the public BAGEL-7B-MoT checkpoint (\texttt{ema.safetensors}) in bf16 for generation, understanding, and image editing, without fine-tuning. The default maximum editing budget is $K=4$, except when explicitly varied in the iteration study, and the random seed is 42.}

\rev{\paragraph{Initial generation.} We use BAGEL's default text-to-image settings: \texttt{cfg\_text\_scale}=4.0, \texttt{cfg\_img\_scale}=1.0, \texttt{cfg\_interval}=$[0.4,1.0]$, \texttt{timestep\_shift}=3.0, 50 denoising steps, CFG renormalization mode \texttt{global}, \texttt{cfg\_renorm\_min}=0.0, and $1024\times1024$ output resolution.}

\rev{\paragraph{Image editing.} Edits use \texttt{cfg\_text\_scale}=4.0, \texttt{cfg\_img\_scale}=2.0, \texttt{cfg\_interval}=$[0.0,1.0]$, \texttt{timestep\_shift}=3.0, 50 denoising steps, CFG renormalization mode \texttt{text\_channel}, and \texttt{cfg\_renorm\_min}=0.0. The output retains the input resolution.}

\rev{\paragraph{Thinking and understanding.} Native thinking is off for the BAGEL and T-GUI rows of the main TIIF-Bench comparison and the WISE rows without CoT. In WISE rows labeled w/ CoT, it is enabled for $Image_0$ only, with greedy decoding and at most 1,000 thinking tokens; the verification--editing loop is unchanged. The prompt-reasoning baseline in Table~\ref{tab: trade-off} is evaluated separately on the same BAGEL backbone. For verification, we insert the original prompt into the understanding prompt of Figure~\ref{fig: under_prompt}, use greedy decoding, and generate at most 1,000 new tokens. A response beginning with \texttt{MATCH} ends the loop. A response beginning with \texttt{EDIT\_NEEDED} passes the text after the colon verbatim to the editor. If neither prefix is present, the entire response is used as the editing instruction. After the $K$-th edit, the output is returned without another verification, as in Eq.~\ref{eq:final_output}.}

\rev{\paragraph{Evaluation and baseline results.} TIIF-Bench uses the original TestMini split: 277 prompts, each with a short and long version (554 prompt--length pairs), across 39 dimension types. We use the official yes/no question lists and GPT-4o evaluator; Text uses the official PaddleOCR and Global Normalized Edit Distance (GNED) scripts. The Style, Text, and Real World subsets contain 30, 30, and 25 prompts, respectively. WISE uses the original 1,000 prompts and GPT-4o WiScore evaluation.}

\section{Additional Result}
\label{Appendix: Additional_visual_results}

\begin{figure}[t!]
    \centering
    \includegraphics[width=0.8\textwidth]{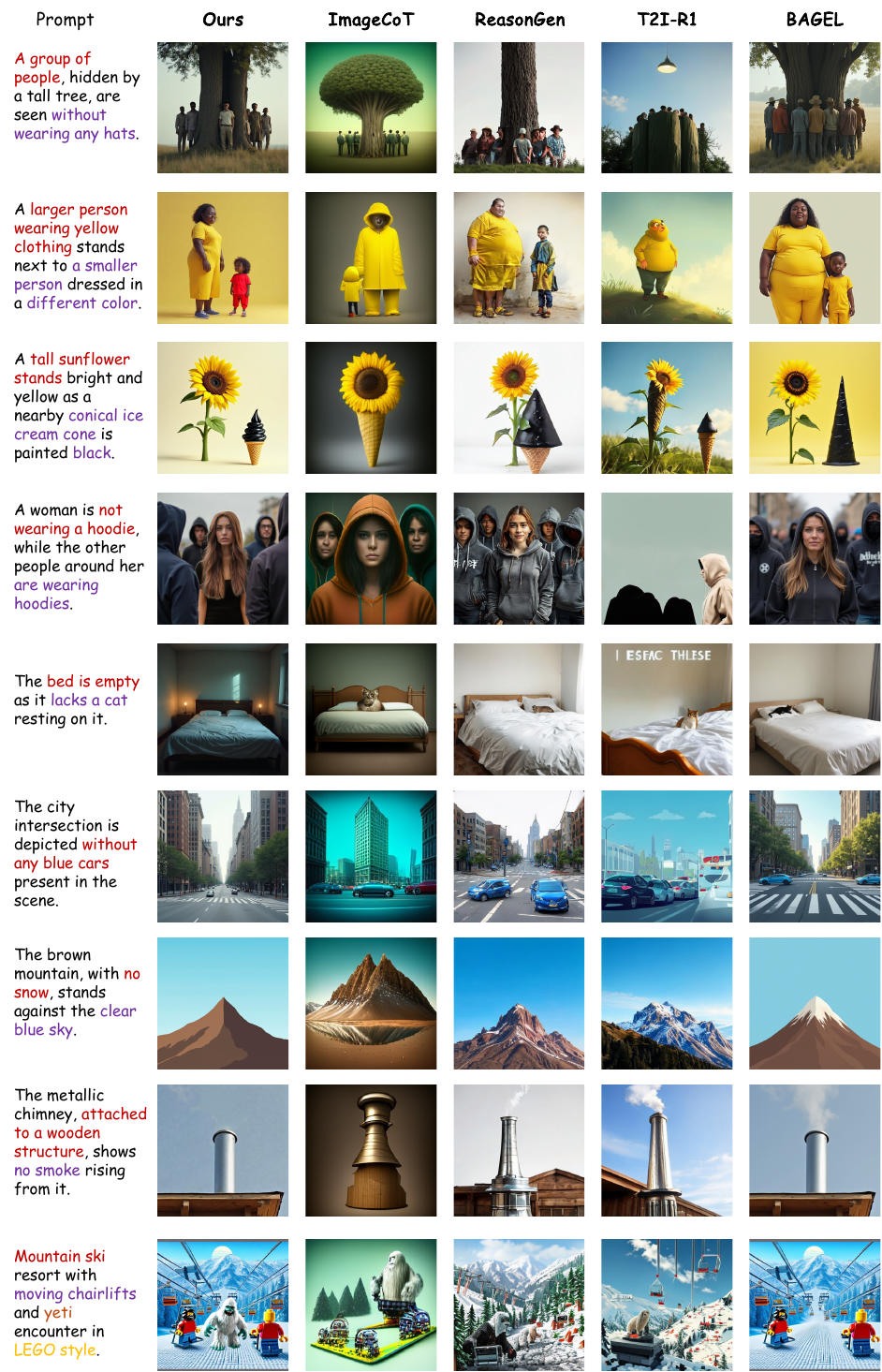}
    \caption{Additional visual comparison with short prompts.}
    \label{fig: additional_visual_comp_1}
\end{figure}

\begin{figure}[t!]
    \centering
    \includegraphics[width=0.75\textwidth,height=0.82\textheight,keepaspectratio]{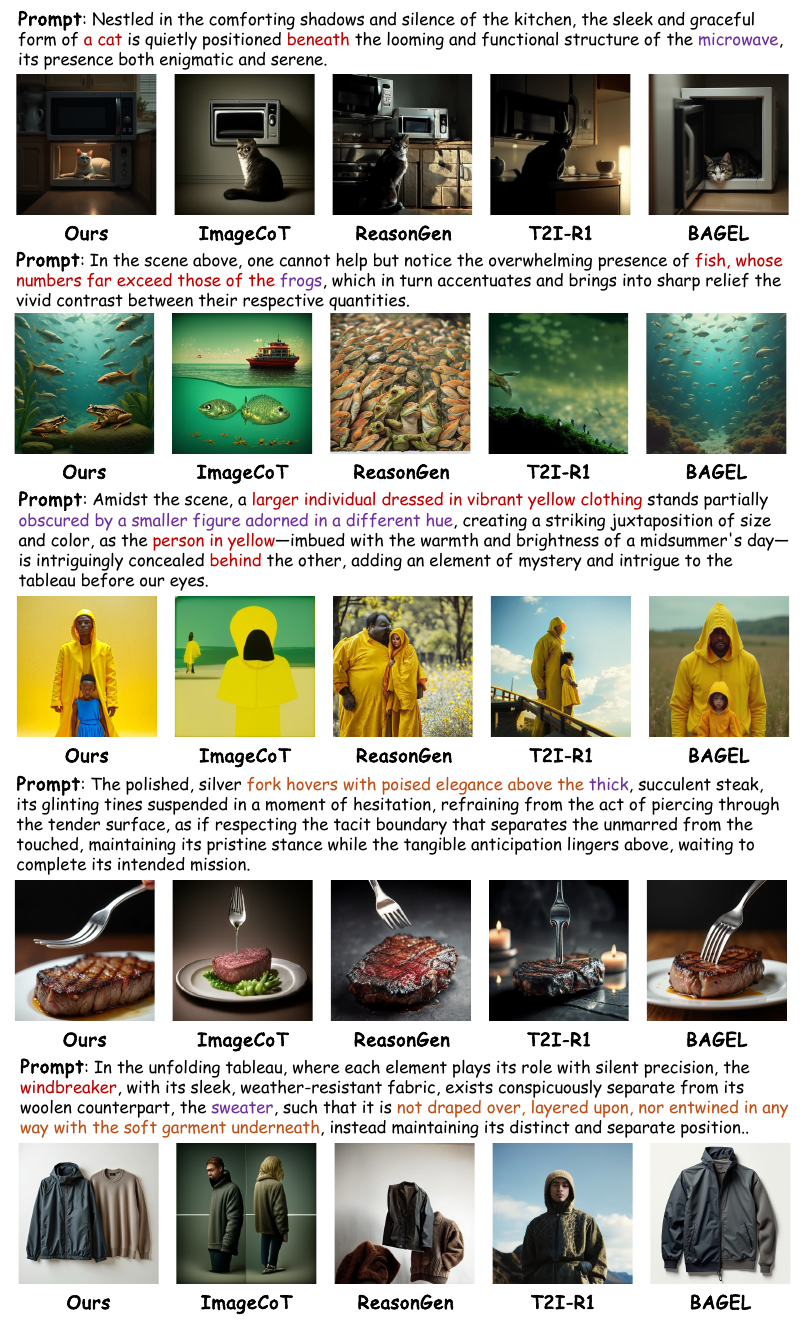}
    \caption{Additional visual comparison with long prompts (group 1).}
    \label{fig: additional_visual_comp_2}
\end{figure}

\begin{figure}[t!]
    \centering
    \includegraphics[width=0.75\textwidth]{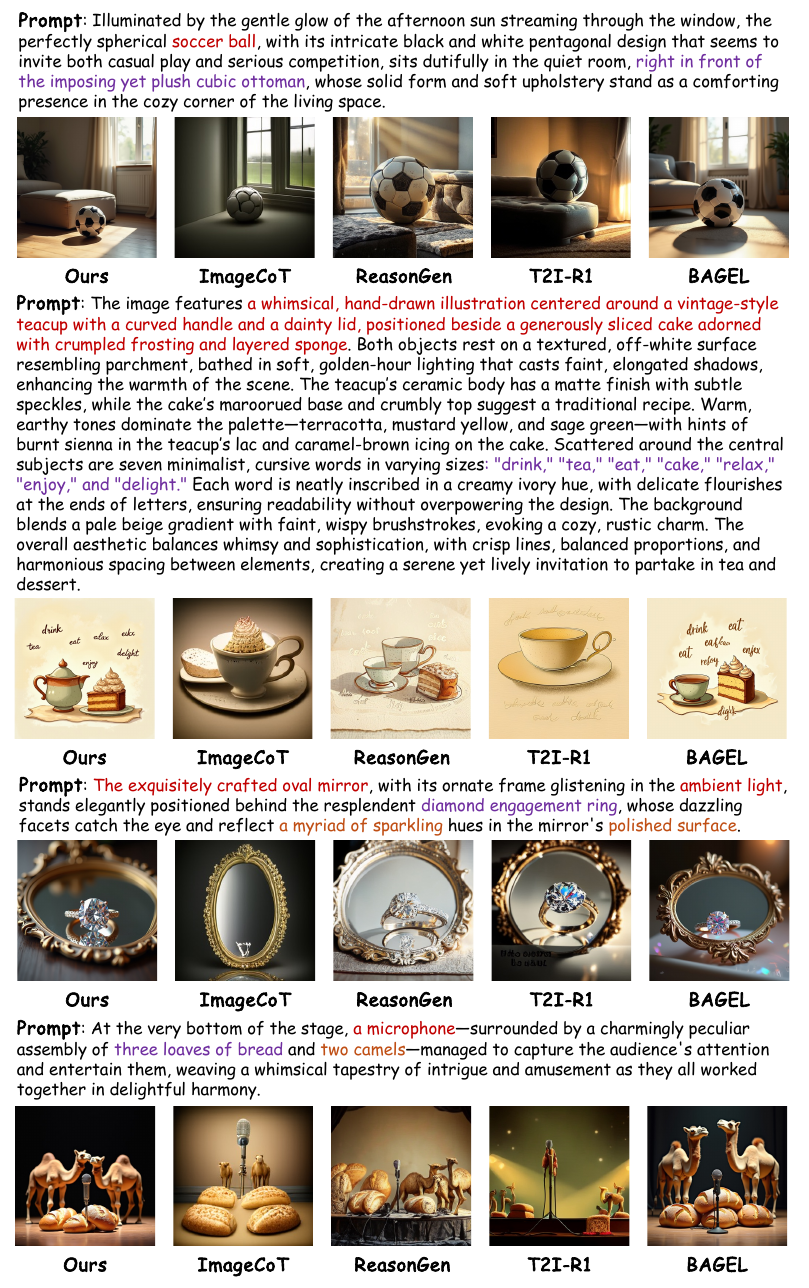}
    \caption{Additional visual comparison with long prompts (group 2).}
    \label{fig: additional_visual_comp_3}
\end{figure}

\subsection{More Visual Comparison}
\label{Appendix: Additional_visual_comparison}

\begin{table}[t!]
\renewcommand{\tabcolsep}{8pt}
\caption{\textbf{Results of T-GUI with multiple UMM backbones.}}
\resizebox{\linewidth}{!}{
\begin{tabular}{lcccccc}
\toprule
\textbf{Model} & \multicolumn{3}{c}{Short Prompt} & \multicolumn{3}{c}{Long Prompt} \\
\midrule
 & Without T-GUI & With T-GUI & Gain & Without T-GUI & With T-GUI & Gain \\
\midrule
Bagel & 68.25 & 69.70 & \textbf{+1.45} & 66.54 & 71.11 & \textbf{+4.57} \\
Janus-Pro & 66.50 & 67.87 & \textbf{+1.37} & 65.02 & 67.42 & \textbf{+2.40} \\
Show-o & 59.72 & 61.03 & \textbf{+1.31} & 58.86 & 62.34 & \textbf{+3.48} \\
\bottomrule
\end{tabular}
}
\label{tab:multi-backbone}
\end{table}

As Table~\ref{tab:multi-backbone} shown, we apply our proposed T-GUI with various UMM base models, including Bagel, Janus-Pro, and Show-o, under both short and long prompts of TIIF benchmark. These results show that T-GUI consistently improves multiple UMM backbones, which provides direct evidence that T-GUI can indeed elicit inference-time self-correction ability across different unified multimodal models, rather than benefiting only from a particular backbone. In particular, the gains become larger in the long-prompt setting, which is consistent with our claim that generation–understanding interaction is especially helpful when prompt following becomes more challenging.
We present more visual comparison in Figure~\ref{fig: additional_visual_comp_1}~\ref{fig: additional_visual_comp_2}~\ref{fig: additional_visual_comp_3}. 

\newpage

\subsection{More Visualization of Reasoning Process}
\label{Appendix: Additional_reasoning_visualization}
\begin{figure}[htbp]
    \centering
    
    \begin{overpic}[width=0.68\textwidth]{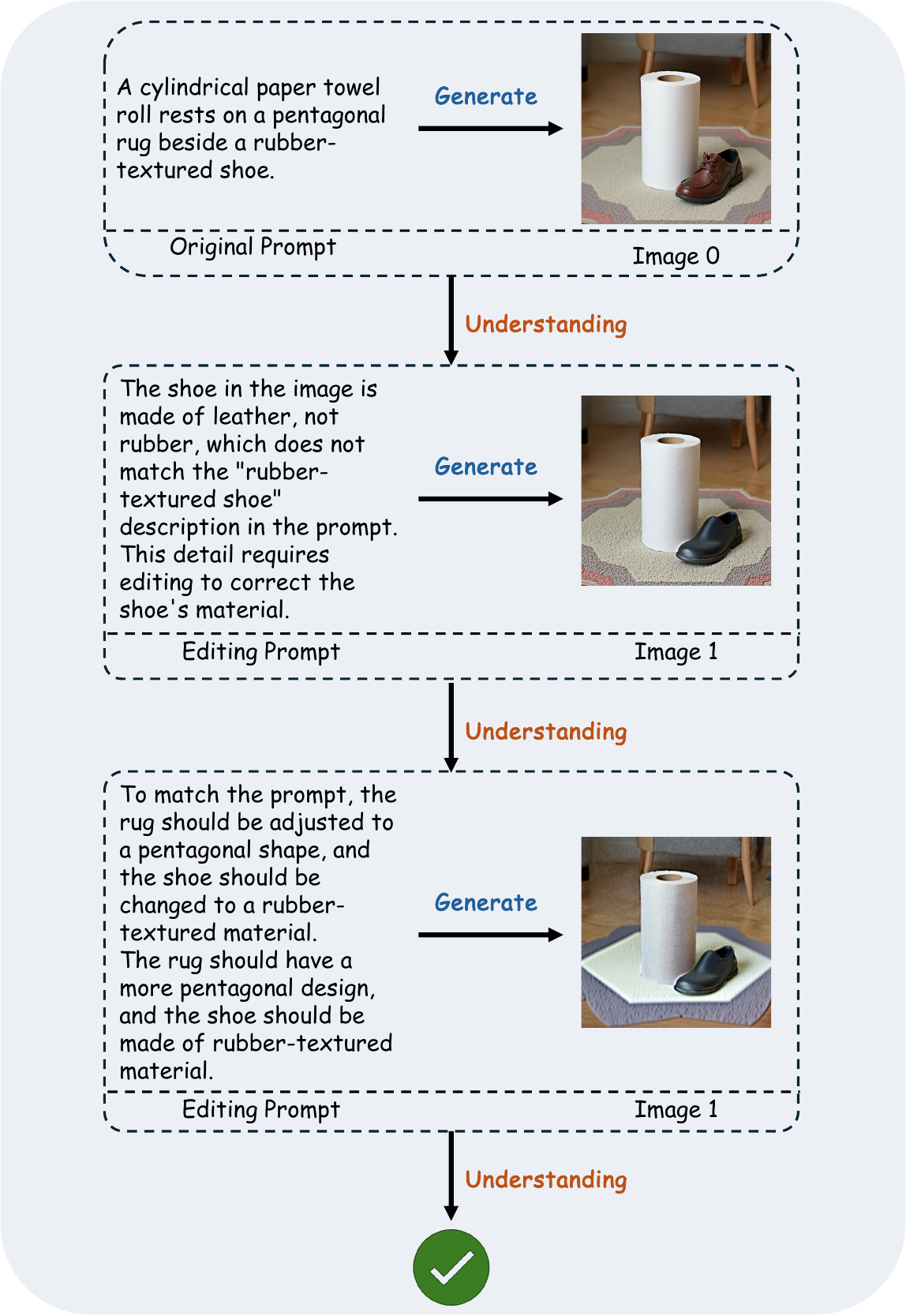}
      \put(48.0,14.3){\makebox(0,0)[lb]{\colorbox{trajectorybg}{\textcolor{black}{\fontsize{8}{9}\selectfont Image 2}}}}
    \end{overpic}
    
    \caption{Additional visualization of the self-correction process (group 1).}
    \label{fig: additional_visual_reason_1}
\end{figure}

\begin{figure}[t!]
    \centering
    \begin{overpic}[width=0.65\textwidth]{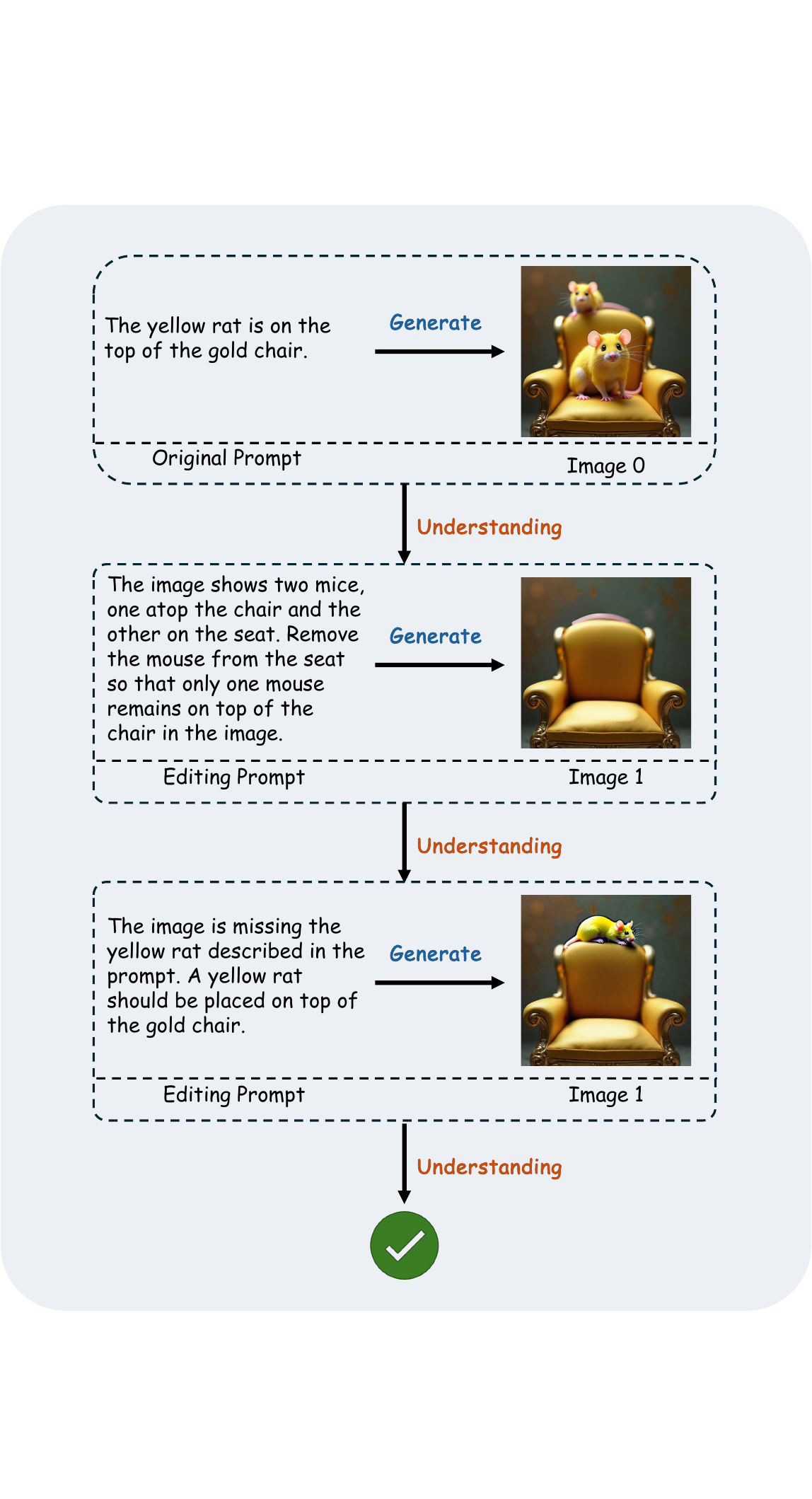}
      \put(37.8,25.9){\makebox(0,0)[lb]{\colorbox{trajectorybg}{\textcolor{black}{\fontsize{8}{9}\selectfont Image 2}}}}
    \end{overpic}
    \caption{Additional visualization of the self-correction process (group 2).}
    \vspace{8pt}
    \label{fig: additional_visual_reason_2}
\end{figure}

\begin{figure}[t!]
    \centering
    \includegraphics[width=0.65\textwidth]{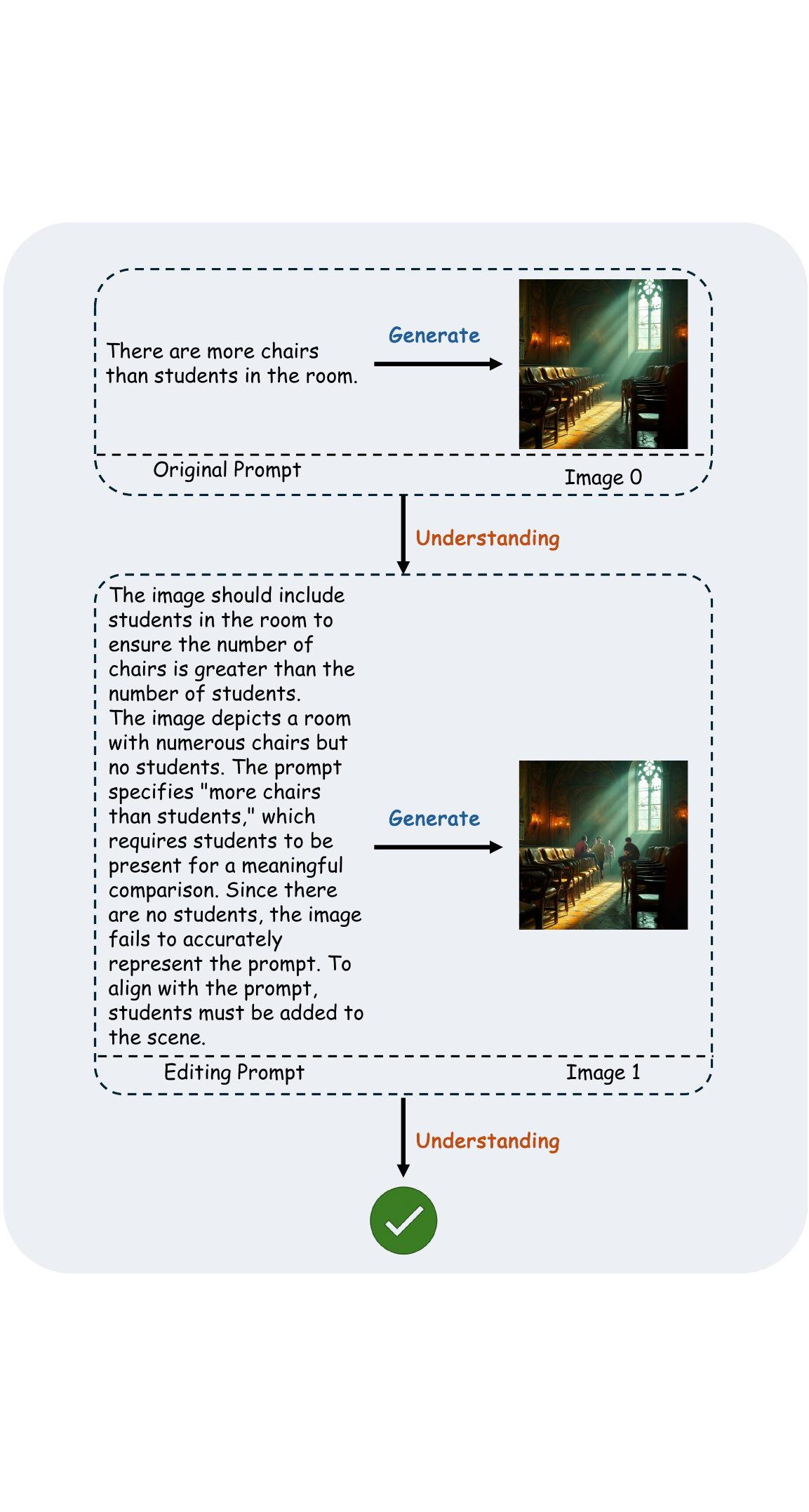}
    \caption{Additional visualization of the self-correction process (group 3).}
    \label{fig: additional_visual_reason_3}
\end{figure}

\begin{figure}[t!]
    \centering
    \includegraphics[width=0.65\textwidth]{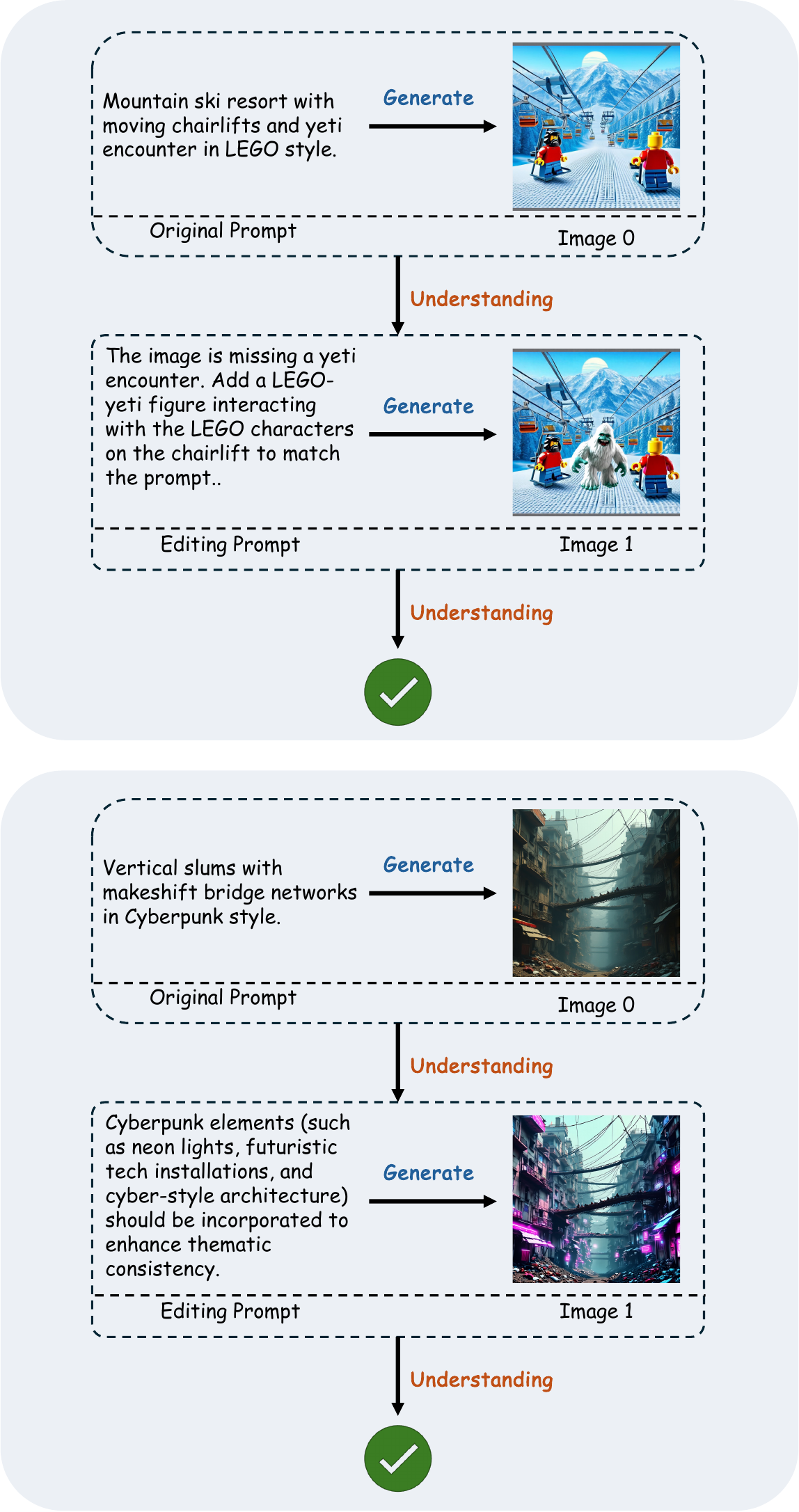}
    \caption{Additional visualization of the self-correction process (group 4).}
    \label{fig: additional_visual_reason_4}
\end{figure}

We show additional visualization of the self-correction process in Figure~\ref{fig: additional_visual_reason_1}~\ref{fig: additional_visual_reason_2}~\ref{fig: additional_visual_reason_3}~\ref{fig: additional_visual_reason_4}.
The visualization illustrates the full self-correction process of our proposed T-GUI, which demonstrates the effectiveness of our T-GUI to guide generation step by step via understanding.

\clearpage
\newpage

\section{Additional Discussion}



\subsection{Failure Case}
\label{app:failure_cases}

\begin{figure}[t!]
    \centering
    \includegraphics[width=\textwidth]{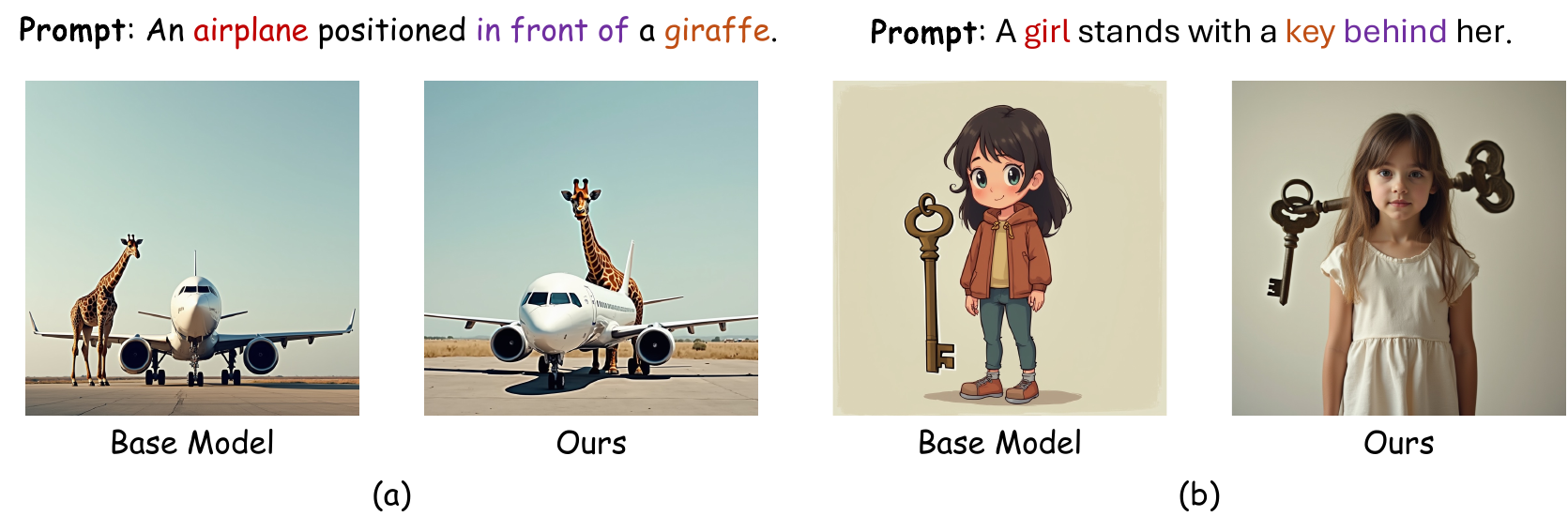}
    \vspace{-18pt}
    \caption{The failure cases.}
    \label{fig: failure_case}
\end{figure}

Our T-GUI demonstrates a significant improvement in the generative capabilities of UMM. However, due to the limitations in the spatial understanding capabilities of the UMM, generating images that depict rare spatial relationships between objects can lead to certain ``inconsistencies". As shown in Figure~\ref{fig: failure_case} \textbf{(a)}, while the T-GUI successfully generates the correct positional relationship between the giraffe and the airplane (\textit{the giraffe is behind the airplane}) compared to the base model, the relative size relationship between the giraffe and the airplane appears inconsistent. \rev{Similarly, as illustrated in Figure~\ref{fig: failure_case} \textbf{(b)}, compared to the base model, T-GUI produces the requested relative position, but the keys behind the little girl remain floating, violating physical plausibility.} 
\rev{Improving spatial understanding and editing may reduce these failures, while their detection remains limited by the same model's verification capabilities. Future work will study stronger verification and preservation of physical consistency.}

\subsection{Trade-off}
\label{app:trade_off}
\begin{table*}[t!]
\renewcommand{\tabcolsep}{30pt}
\caption{
\textbf{Analysis of the Trade-off.} We analyze the trade-off between performance and latency on the long
prompt setting of the TIIF benchmark.
}
\resizebox{\linewidth}{!}{
\begin{tabular}{l|cc}
\toprule
\textbf{Model} & \textbf{Performance} $\uparrow$ & \textbf{Average Latency (s)} $\downarrow$ \\
\midrule
Verify-based Reasoning & 67.18 & 391.92 \\
Prompt Reasoning & \underline{67.90} & \textbf{108.23} \\
Our T-GUI & \textbf{71.11} & \underline{157.61} \\
\bottomrule
\end{tabular}
}
\label{tab: trade-off}
\end{table*}
As shown in Table~\ref{tab: trade-off}, we analyze the trade-off between performance and latency across the verification-based framework, the prompt-based framework, and our proposed T-GUI. 
To ensure a fair comparison, the analysis is conducted using the same base model (BAGEL~\citep{deng2025emerging}). 
Our T-GUI achieves the best performance with lower latency than the verification-based framework, although its latency is slightly higher than that of the prompt-based framework.
These results demonstrate the overall advantage of T-GUI in balancing performance and latency.

\section{User Study}
\label{Appendix: User_study}

\begin{figure}[t!]
    \centering
    \includegraphics[width=\textwidth]{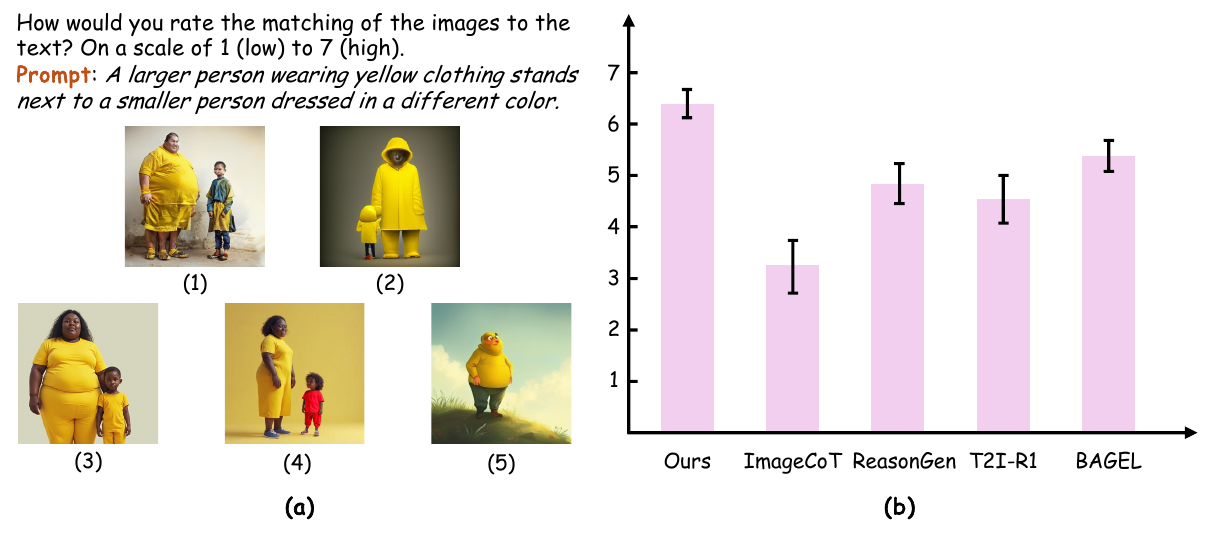}
    \caption{The user study. \textbf{(a)} An example of the question. \textbf{(b)} The results of the user study.}
    \label{fig: user_study}
\end{figure}

To further evaluate the text-to-image generation capabilities for baselines and our T-GUI, we conducted a user study focused on participants’ subjective assessments of generation quality and the alignment between generated images and the given prompts.

\noindent \textbf{Participants.}
The study involved 20 participants, 35\% of whom were aged 18–25 and 65\% aged 25–35. The gender distribution was 60\% male and 40\% female. Notably, 85\% of the participants reported prior experience with AI technologies.

\noindent \textbf{Task and Measurement.}
As illustrated in Figure~\ref{fig: user_study} \textbf{(a)}, participants were asked to rate a series of images on a 7-point Likert scale, evaluating both visual quality and the depth of generation relative to the associated input prompts. Images generated by various models were presented in randomized order to control for order effects.

\noindent \textbf{Results.}
As shown in Figure~\ref{fig: user_study} \textbf{(b)}, participants consistently rated the quality and generation accuracy of images generated by our T-GUI framework, with the highest mean scores across all comparisons. This indicates a significant performance advantage over baseline models. Furthermore, T-GUI demonstrated relatively low variance in ratings, underscoring its consistency and reliability in text-to-image generation.
Taken together, the combination of high average scores and reduced variability suggests that our proposed T-GUI not only produces more accurate and contextually appropriate images but also does so more consistently across diverse prompts.

\end{document}